\documentclass[journal]{IEEEtran}
\usepackage{preamble, my-notations}

\title{Learning to Gridize: Segment Physical World by Wireless Communication Channel}

\author{
Juntao~Wang,
Feng~Yin,~\IEEEmembership{Senior Member,~IEEE,}
Tian~Ding,
Tsung-Hui~Chang,~\IEEEmembership{Fellow,~IEEE,}
Zhi-Quan~Luo,~\IEEEmembership{Fellow,~IEEE,}
and~Qi~Yan
\thanks{The work was supported by the National Key Research and Development Program under grant 2024YFA1014201, the National Key Research and Development Program under grant 2022YFA1003900, the Guangdong Major Project of Basic and Applied Basic Research (No.2023B0303000001), and the Guangdong Provincial Key Laboratory of Big Data Computing. \textit{(Corresponding author: Feng Yin.)}}
\thanks{Juntao Wang and Zhi-Quan Luo are with the Shenzhen Research Institute of Big Data and also with the School of Science and Engineering, The Chinese University of Hong Kong (Shenzhen), Guangdong 518172, China (email: juntaowang@link.cuhk.edu.cn; luozq@cuhk.edu.cn).}
\thanks{Feng Yin is with School of Science and Engineering, The Chinese University of Hong Kong (Shenzhen), Guangdong 518172, China (email: yinfeng@cuhk.edu.cn).}
\thanks{Tsung-Hui Chang is with the School of Artificial Intelligence, The Chinese University of Hong Kong, Guangdong 518172, China, and also with Shenzhen Research Institute of Big Data, Guangdong 518172, China (email: tsunghui.chang@ieee.org).}
\thanks{Tian Ding is with Shenzhen Research Institute of Big Data, Guangdong 518172, China (email: dingtian@sribd.cn).}
\thanks{Qi Yan is with the Networking and User Experience Laboratory, Huawei Technologies, Guangdong 518129, China (email: yanqi1@huawei.com).}
}

\begin{document}

\ifCLASSOPTIONpeerreview
\IEEEpeerreviewmaketitle
\else
\maketitle
\fi

\setcounter{page}{1}

\begin{abstract}
    Gridization, the process of partitioning space into grids where users share similar channel characteristics, serves as a fundamental prerequisite for efficient large-scale network optimization.
    However, existing methods like Geographical or Beam Space Gridization (GSG or BSG) are limited by reliance on unavailable location data or the flawed assumption that similar signal strengths imply similar channel properties.
    We propose Channel Space Gridization (CSG), a pioneering framework that unifies channel estimation and gridization for the first time.
    Formulated as a joint optimization problem, CSG uses only beam-level reference signal received power (RSRP) to estimate Channel Angle Power Spectra (CAPS) and partition samples into grids with homogeneous channel characteristics.
    To perform CSG, we develop the CSG Autoencoder (CSG-AE), featuring a trainable RSRP-to-CAPS encoder, a learnable sparse codebook quantizer, and a physics-informed decoder based on the Localized Statistical Channel Model.
    On recognizing the limitations of naive training scheme, we propose a novel Pretraining-Initialization-Detached-Asynchronous (PIDA) training scheme for CSG-AE, ensuring stable and effective training by systematically addressing the common pitfalls of the naive training paradigm.
    Evaluations reveal that CSG-AE excels in CAPS estimation accuracy and clustering quality on synthetic data. On real-world datasets, it reduces Active Mean Absolute Error (MAE) by 30\% and Overall MAE by 65\% on RSRP prediction accuracy compared to salient baselines using the same data, while improving channel consistency, cluster sizes balance, and active ratio, advancing the development of gridization for large-scale network optimization.
\end{abstract}


\section{Introduction}\label{sec:introduction}

\IEEEPARstart{T}{he} proliferation of wireless devices and relentless growth of mobile data traffic have catalyzed a data-driven revolution in network optimization, with massive network data enabling novel optimization paradigms~\cite{luo2023srcon}. 
Conventional approaches have largely focused on individual user- or sample-level optimization, treating each mobile user as a single optimization instance \cite{liu2024survey}. While this granular approach offers intuitive simplicity, it proves increasingly inadequate in real-world deployments confronted with unprecedented sample volumes and unstable user dynamics.

First, the data deluge makes optimization at sample-level granularity computationally expensive for networks serving millions of devices. This necessitates the extraction of higher-level features as small-scale proxies for efficient optimization. Second, traditional user- or sample-level approaches frequently struggle to adapt to changing user populations or channel conditions, resulting in brittle or oscillatory optimization policies in dynamic scenarios \cite{tao2024parallel}. Take the sum-utility maximization problem as an example, let $u_i(\Psi)$ be the utility function of the $i$-th user equipment (UE) under network configuration $\Psi$ with constraints $\mathcal{C}(\Psi)$, there is a mismatch between the optimal configurations for different UE populations $\set{I}_1 \neq \set{I}_2$:
\begin{equation}\label{eq: sum_utility_max}
    \Psi_{1}^{*} \triangleq \argmax_{\Psi \in \mathcal{C}(\Psi)} \sum_{i \in \set{I}_1} u_i(\Psi) \, \ne \, \Psi_{2}^{*} \triangleq \argmax_{\Psi \in \mathcal{C}(\Psi)} \sum_{i \in \set{I}_2} u_i(\Psi).
\end{equation}
This fundamental mismatch arises from the sample-level definition of utility functions, which lack consistency across varying UE sets. Consequently, there is an urgent need to develop more robust and static representations of UEs that can be effectively reused across different UE populations.

\begin{figure*}[tb]
    \centering
    \subfloat[]{%
        \includegraphics[height=0.203\linewidth]{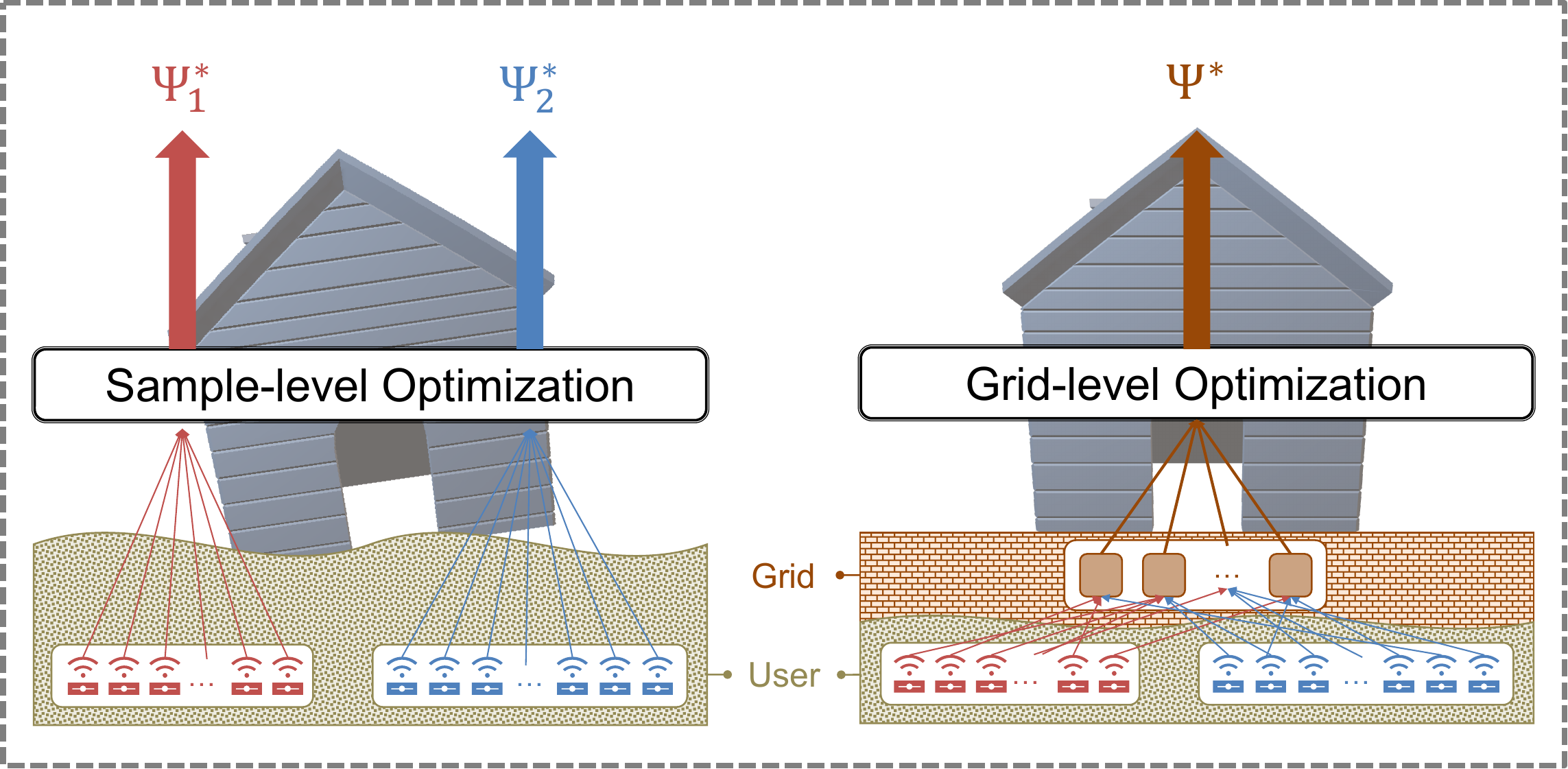}%
        \label{fig:sample_grid_level_opt}
    }
    \hfill
    \subfloat[]{%
        \includegraphics[height=0.203\linewidth]{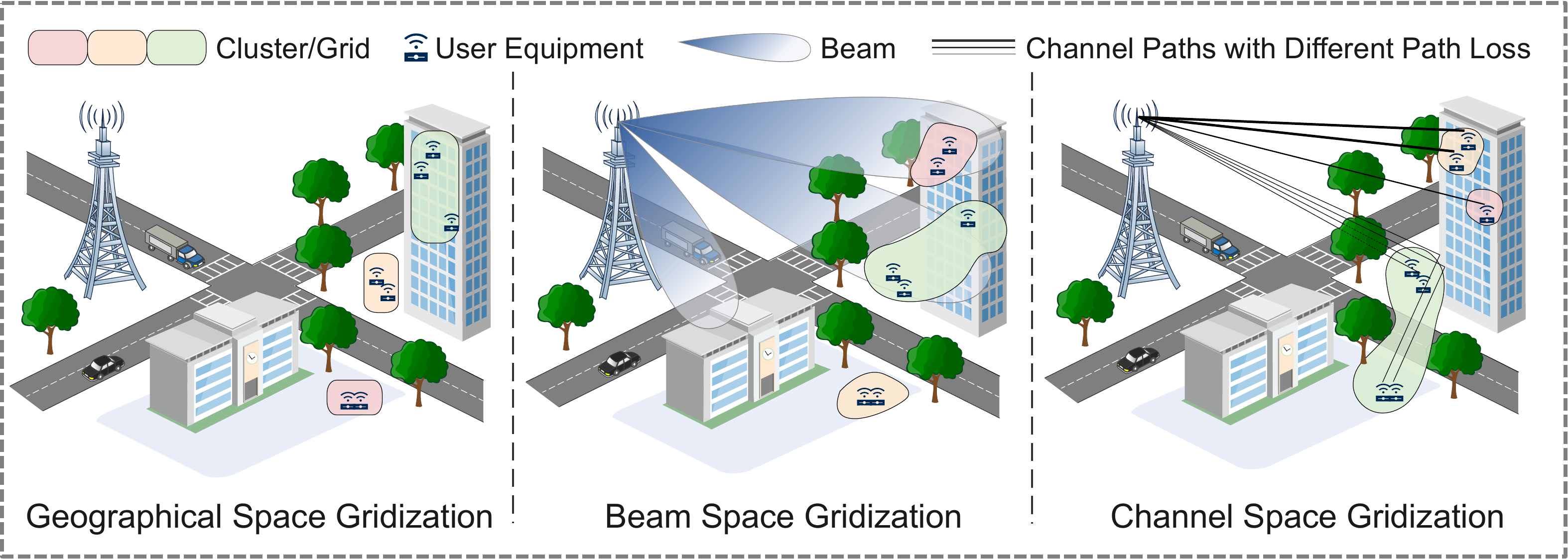}%
        \label{fig:gridization}
    }
    \caption{(a) Sample-level \vs~grid-level network optimization. (b) Different gridization methods in wireless communication networks.}
    \label{fig:intro_long}
    \vspace{-1em}
\end{figure*}

Gridization, a cornerstone in data-driven network optimization, addresses these challenges by transforming continuous coverage areas into discrete operational units called grids. The name of gridization originates from network management practices, which 
facilitates efficient coarse-grained management through spatial partitioning. In the context of data-driven optimization, it facilitates grid-level performance modeling by clustering data points with similar characteristics. The motivation for gridization rests on two key pillars:
\begin{itemize}[leftmargin=*]
    \item \textbf{Efficiency---Data Compression:} The massive scale of measurement data necessitates efficient representation. Gridization accomplishes this by clustering data, significantly reducing storage and computational demands for subsequent processing. This data compression is vital for large-scale network optimization, enabling efficient and scalable optimization while preserving the essential characteristics of the original dataset.
    \item \textbf{Robustness---Environmental Modeling and Reusability:} Gridization aggregates inherently random measurements into meaningful clusters, forming a reusable ``digital asset'' that characterizes the communication environment, where each cluster of data corresponds to a subregion of the network coverage area. As the environment remains stable over short periods, gridization results provide robust and reusable optimization instances that support grid-level optimization. When physical surroundings change significantly, only minimal new data are needed to update the gridization, as previous results serve as accumulated 
    environmental knowledge, reducing the effort and cost of model refinement.
\end{itemize}

Formally, gridization generates a set of grids $\set{K}$, where $|\set{K}| \ll |\set{I}|$. An ideal grid serves as a container for UEs sharing similar communication characteristics, \eg, channel properties, thereby ensuring that UEs within the same grid experience consistent network performance changes during configuration adjustments. This enables the definition of a grid-level utility function $\tilde{u}_k(\Psi)$, which reformulates the sum-utility maximization problem in \eqref{eq: sum_utility_max} as:
\begin{equation}\label{eq: sum_utility_max_grid}
    \max_{\Psi \in \mathcal{C}(\Psi)} \sum_{k \in \set{K}} \tilde{u}_k(\Psi).
\end{equation}
The grid-level utility function $\tilde{u}_k(\Psi)$ approximates the aggregated utility across all user equipments within grid $k$. This formulation enables a two-stage optimization procedure: first, UEs are assigned to corresponding grids; subsequently, the network configuration is optimized for each grid, regardless of how the UE set $\set{I}$ evolves over time. Figure~\ref{fig:sample_grid_level_opt} abstractly contrasts sample-level and grid-level optimization.

Existing gridization techniques predominantly follow two paradigms, each with intrinsic limitations:

\paragraph{\textbf{Geographical Space Gridization (GSG)}} 
GSG partitions the coverage area into geographical grids by clustering UEs based on their spatial proximity, utilizing location information \cite{ning2022multi, zhang2024lscm}. However, location data is frequently unavailable or unreliable due to privacy concerns or UE configuration settings \cite{yin2020fedloc, luo2023srcon}. To overcome this limitation, drive tests (DTs) are typically employed, involving extensive measurements with specialized equipment across various locations to simulate multiple UEs. Nevertheless, DT data suffers from limited scale and high collection costs, rendering geographical gridization impractical for large-scale deployments.

\paragraph{\textbf{Beam Space Gridization (BSG)}} 
BSG as a more economical alternative utilizes massive measurement report (MR) data from the network management system, which includes beam-level reference signal received power (RSRP) measurements across all UEs \cite{3gpp38133}. This approach operates in beam space, which uses beam-level RSRP measurements as coordinates, and clusters measurements into beam space grids \cite{ying2022space}. However, critical flaw persists: RSRP similarity does not guarantee similar channel, as different multipath components can produce identical received power through constructive/destructive interference. This leads to grid misclassification where UEs with different channel properties but comparable RSRP values get grouped together, ultimately degrading optimization outcomes.

To address the limitations of existing methods, we introduce \emph{Channel Space Gridization}, a novel approach that represents a paradigm shift in gridization:

\paragraph{\textbf{Channel Space Gridization (CSG)}} 
CSG integrates channel modeling and gridization into a unified framework for the first time. Unlike GSG, which depends on expensive location information, CSG leverages extensive RSRP samples from MR data to jointly estimate channels and partition samples into grids based on these estimates. Compared to BSG, CSG encourages RSRP samples with similar channel characteristics to be clustered together, resulting in a more accurate and intrinsically meaningful gridization process.

Recent advances in deep learning, particularly in neural discrete representation learning, have led to powerful methods such as vector quantized variational autoencoders (VQ-VAE) \cite{van2017vqvae}. These techniques integrate discrete latent spaces within neural architectures, enabling effective unsupervised clustering and discrete representation learning, with demonstrated success in computer vision \cite{ramesh2021zero}, speech processing \cite{dhariwal2020jukebox}, multi-modal learning \cite{qu2025tokenflow} and channel coding \cite{nemati2023vq}. Motivated by these advances, this work is, to the best of our knowledge, the first to adapt neural discrete representation learning to the gridization problem in wireless communications. This novel approach enables joint channel modeling and gridization, establishing a new link between advanced machine learning techniques and practical wireless system applications. The key contributions of this paper are as follows:
\begin{itemize}[leftmargin=*]
    \item \textbf{Novel Gridization Framework}: We introduce \emph{Channel Space Gridization} (CSG), a groundbreaking paradigm that unifies channel modeling and spatial partitioning. Unlike traditional GSG or BSG, CSG operates independently of location data while ensuring consistent channel characteristics within grids. We formulate CSG as a joint optimization problem combining channel angle power spectra (CAPS) estimation with sample clustering. The framework incorporates essential constraints on CAPS sparsity and zero-mean perturbations of dominant paths, encouraging long-term stable channel representations within each grid.
    \item \textbf{Deep Learning-based Solution}: We develop the \emph{Channel Space Gridization Autoencoder} (CSG-AE), a deep learning solution that efficiently addresses the CSG problem. The architecture comprises three key components: an encoder that transforms RSRP samples into CAPS representations, a quantizer that maps CAPS to grid centers, and a decoder that reconstructs RSRP using the localized statistical channel model (LSCM) \cite{zhang2024lscm}. This design enables joint optimization of grid partitioning and channel estimation without requiring location information. The model's capability to predict RSRP across varying beam patterns without retraining significantly enhances its practical value for network optimization, offering a scalable and cost-effective solution for large-scale deployments.
    \item \textbf{Improved Training Scheme}: We propose \textbf{PIDA}, a tailored training scheme for CSG-AE that addresses critical challenges in naive training: codebook collapse, update hysteresis, and gradient direction conflicts. PIDA employs encoder \textbf{P}retraining and \kmeans~codebook \textbf{I}nitialization for mitigating codebook collapse, followed by \textbf{D}etached and \textbf{A}synchronous updates to resolve assignment hysteresis and balance competing effects between reconstruction and quantization objectives, which significantly improves training stability and model performance.
    \item \textbf{Comprehensive Experimental Validation}: We demonstrate CSG's superior performance compared to conventional methods through extensive evaluation on both synthetic and real-world datasets. On synthetic data, CSG-AE achieves exceptional CAPS estimation accuracy while maintaining strong clustering quality. On real-world data, our method delivers substantial improvements in RSRP prediction accuracy, enhanced channel consistency within grids, balanced cluster sizes, and top-tier active ratio, all accomplished without location data.
\end{itemize}

The remainder of this paper is organized as follows. Section \ref{sec:preliminary} reviews the LSCM and discusses the limitations of current gridization methods. Section \ref{sec:problem_formulation} formulates the CSG problem. Section \ref{sec:proposed_method} presents the CSG-AE framework, the motivation for adopting the learning-based approach, the proposed training scheme, and the procedure for RSRP prediction under varying beam patterns. Section \ref{sec:experimental_results} provides comprehensive experimental results comparing CSG-AE with baseline methods. Section~\ref{sec:conclusion} concludes the paper and discusses future works.

Throughout this paper, we adopt the following notational conventions: \underline{\textbf{bold}} letters (\eg, $\vec{a}$, $\mat{A}$) denote vectors and matrices, \underline{regular} letters (\eg, $a$, $A$) represent scalars, \underline{\textit{italic}} type (\eg, $N$, $M$, $x$, $\vec{x}$, $\mat{X}$) indicates deterministic variables, \underline{\textrm{roman}} type (\eg, $\rv{x}$, $\rvec{x}$, $\rmat{X}$) designates random variables, and \underline{\textsf{sans-serif}} font (\eg, $\set{I}\triangleq\{1, \dots, I\}$, $\set{X_{(K)}}\triangleq\{\vec{x}_{(1)}, \dots, \vec{x}_{(K)}\}$) represents sets. Additionally, $[\cdot]_{m,n}$ denotes the $(m,n)$-th entry of a matrix, $[\cdot]_{m}$ denotes the $m$-th entry of a vector, $(\cdot)^{\top}$ and $(\cdot)^{H}$ represent transpose and Hermitian transpose operations respectively, $|\cdot|$ indicates the cardinality of a set, $\odot$ denotes the Hadamard (element-wise) product, and $\jmath$ represents the imaginary unit. Any exceptions will be explicitly noted in the context case by case.

\section{Preliminary}\label{sec:preliminary}

In this section, we introduce the system model, local statistical channel model (LSCM)~\cite{zhang2024lscm}, as a preliminary and discuss the limitations of existing methods.

\subsection{Localized Statistical Channel Model}\label{subsec:LSCM}

\begin{figure}[tb]
    \centering
    \subfloat[]{%
        \includegraphics[height=0.47\linewidth]{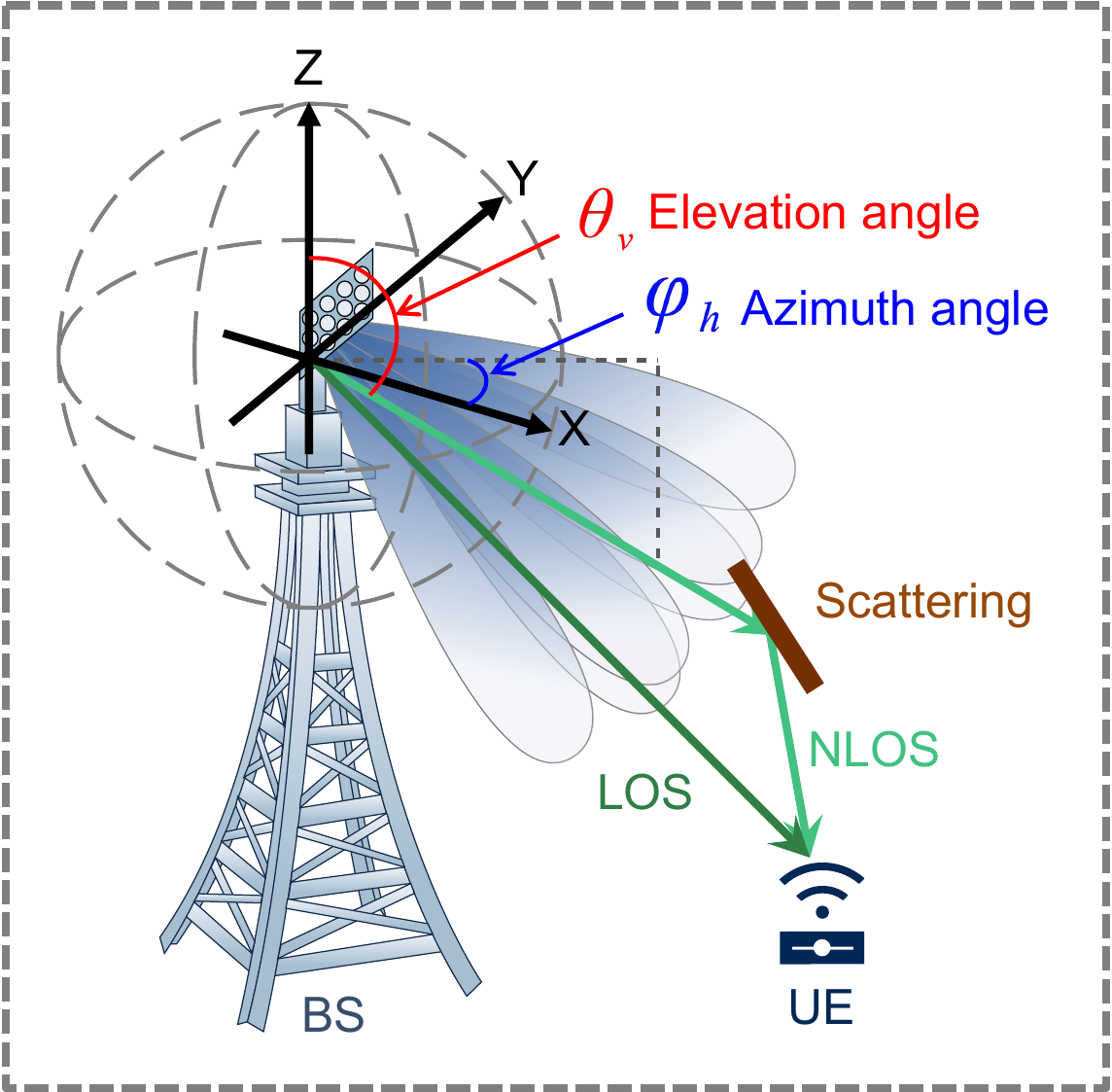}%
        \label{fig:angular_space}
    }
    \hfill
    \subfloat[]{%
        \includegraphics[height=0.47\linewidth]{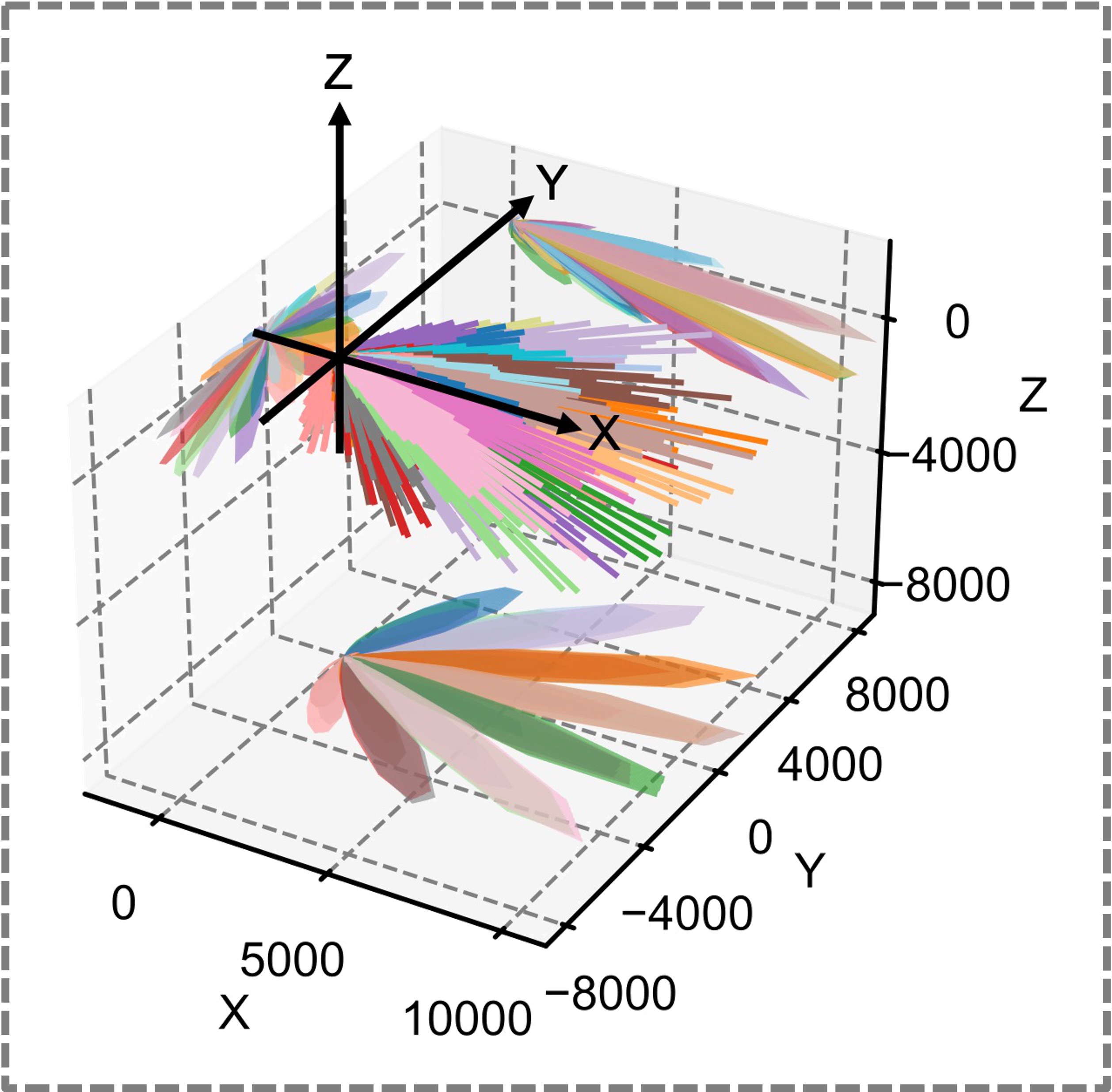}%
        \label{fig:SRS_matrixA}
    }
    \caption{(a) Discrete angular space. (b) Visualization of a beam pattern matrix $\mat{A}$. Specifically, for the entry $[\mat{A}]_{m,n}$ corresponding to the power gain of the $m$-th beam in the $n$-th spatial angle $(\theta_{v}, \varphi_{h})$, we represent it as a vector in the $m$-th color, originating from the origin and terminating at the point $(x, y, z) = \left([\mat{A}]_{m,n}\sin\theta_{v} \cos\varphi_{h}, [\mat{A}]_{m,n}\sin\theta_{v} \sin\varphi_{h}, [\mat{A}]_{m,n}\cos\theta_{v}\right)$. On the XY, XZ, and YZ planes, the vectors are projected onto these planes respectively for better visualization.}
    \label{fig:angular_space_and_SRS_matrixA}
    \vspace{-1em}
\end{figure}

The LSCM considers a downlink channel model, where the BS is equipped with a uniform rectangular array (URA) comprising $N_T = N_y \times N_z$ transmit antennas, and the UE has one receive antenna.
The antenna panel is positioned in the Y-Z plane, with its $(1,1)$-th antenna element as the coordinate origin and its normal direction aligned along the positive X-axis. 
The free space is discretized into $N_V$ elevation angles $\{\theta_{v}\}_{v=1}^{N_V}$ and $N_H$ azimuth angles $\{\varphi_{h}\}_{h=1}^{N_H}$, forming a discrete angular space with $N = N_V \times N_H$ distinct spatial angles, as depicted in Fig.~\ref{fig:angular_space}.

To detect the channel paths, the BS generates $M$ $(M \ll N)$ beams using the precoding matrix $\mat{B} \in \C^{N_T \times M}$ to sweep the space and transmit reference signals to the UEs. 
In the downlink of 5G system, such reference signals are synchronization signal block (SSB) or channel state information-reference signal (CSI-RS) \cite{cox2020introduction}.
The UE reports the beam-level RSRP to the BS, denoted by $\rvec{RSRP} \in \R^M$, which is given by $ \rvec{RSRP} = P|\mat{B}^{H} \rvec{h}|^2$, where $|\cdot|^2$ takes the element-wise absolute value and square, $P$ is the transmit power, and $\rvec{h} \in \C^{N_T}$ is the channel impulse response (CIR) vector between the BS and UE \cite{tse2005fundamentals}.

The LSCM represents the CIR similarly to the model described in 3GPP TR 38.901 \cite{3gpp38901}, which is given by:
\begin{equation}\label{eq: CIR}
    \rvec{h} = \sum_{v=1}^{N_V} \sum_{h=1}^{N_H} \sqrt{\rv{x}_{v, h}} g_{v, h} \vec{s}_{v, h} \odot \rvec{w}_{v,h}.
\end{equation}
It focuses on the angle of departure (AoD) and channel gain from the BS to UE, while ignoring the angle of arrival (AoA) and delay effects. In \eqref{eq: CIR}, $\rv{x}_{v, h} \ge 0$ denotes the channel gain for paths with AoD in the spatial angle $(\theta_{v}, \varphi_{h})$, and is zero if no path exists. $g_{v, h}$ denotes the antenna gain for the spatial angle $(\theta_{v}, \varphi_{h})$. $\vec{s}_{v, h} \in \C^{N_T}$ is the array steering vector corresponding to the spatial angle $(\theta_{v}, \varphi_{h})$, with entry corresponding to the $(y,z)$-th antenna element given by $\exp{\left[-\jmath \frac{2\pi}{\lambda} \left(d_y y \sin\theta_{v} \sin\varphi_{h} + d_z z \cos\theta_{v} \right)\right]}$. $d_y$ and $d_z$ are the antenna spacing in the $y$ and $z$ directions, respectively, and $\lambda$ is the wavelength. $\rvec{w}_{v,h} \in \C^{N_T}$ is the random phase error vector, with entry corresponding to the $(y,z)$-th antenna element given by $\exp{\left[-\jmath \left(\rv{w}_{v, h} + \rv{w}_{y, z} \right)\right]}$. $\rv{w}_{v, h}$ and $\rv{w}_{y, z}$ are the random phase errors for the spatial angle $(\theta_{v}, \varphi_{h})$ and the $(y,z)$-th antenna element, respectively.

Assuming that $\rv{w}_{v, h}$, $\rv{w}_{y, z}$, and $\rv{x}_{v, h}$ are independent, with $\rv{w}_{v, h} \sim \Uniform{-\pi}{\pi}$ and $\rv{w}_{y, z} \sim \Normal{0}{\sigma^2}$, it has been proved in \cite{ning2022multi, zhang2024lscm} that taking the expectation over random phase errors yields:
\begin{equation}\label{eq: LSCM}
    \rvec{y} \triangleq \E_{\rv{w}_{v, h}, \rv{w}_{y, z}}[\rvec{RSRP}] = \mat{A}(\Psi)\rvec{x}
\end{equation}
where $\mat{A}(\Psi) \in \R^{M \times N}$ is a matrix only determined by the BS-side antenna parameters $\Psi = \left\{P, \mat{B}, d_z, d_y, \lambda \right\}$, and
\begin{equation}\label{eq: channel_APS}
    \begin{aligned}
        \rvec{x} = &\left[ \rv{x}_{1,1}, \rv{x}_{1,2}, \dots, \rv{x}_{1,N_H}, \rv{x}_{2,1}, \dots \right.\\
                           &\left. \rv{x}_{2,N_H}, \dots, \rv{x}_{N_V,1}, \dots \rv{x}_{N_V, N_H} \right]^{\top} \in \R^{N}_{\ge 0}.
    \end{aligned}
\end{equation}

\begin{remark}
    For the detailed expression of $\mat{A}(\Psi)$ and proof of \eqref{eq: LSCM}, please refer to the Appendix of \cite{zhang2024lscm}. The matrix $\mat{A}(\Psi)$ can be computed explicitly using the expression in \cite{zhang2024lscm}, once the parameters $\Psi$ are specified. For notational conciseness, we will henceforth denote $\mat{A}(\Psi)$ simply as $\mat{A}$.
\end{remark}

The coefficient matrix $\mat{A}$ serves as a beam pattern matrix, with each entry $[\mat{A}]_{m,n}$ representing the power gain of the $m$-th beam in the $n$-th spatial angle. This matrix characterizes power distribution across the angular domain, as shown in Fig.~\ref{fig:SRS_matrixA}. The vector $\rvec{x}$ denotes the channel angular power spectrum (CAPS), with each entry indicating channel gain at a spatial angle, thereby capturing the multi-path structure of the propagation environment. Since channel paths are sparsely distributed in angular space, the CAPS $\rvec{x}$ exhibits sparsity.

\begin{remark}
    The LSCM offers significant practical utility in network optimization \cite{luo2023srcon, zhang2024lscm}---once the CAPS $\rvec{x}$ is estimated, regardless of how the network parameters $\Psi$ are tuned, we can compute the corresponding $\mat{A}$ and predict the resulting RSRP by \eqref{eq: LSCM}. This provides a powerful tool for network optimization, as it allows us to find the optimal configuration for network without the need for time-consuming simulations.
\end{remark}

\subsection{Existing Methods and Limitations}\label{subsec:existing_methods}

Previous work \cite{zhang2024lscm,ning2022multi} proposed a geographical-based approach for CAPS estimation, operating under the assumption that RSRP samples from proximate locations exhibit similar CAPS. Their methodology comprises three key steps:
\begin{enumerate}[leftmargin=*]
    \item \textbf{Drive-Test Data Collection}: Conduct drive tests (DTs) to gather RSRP measurements and their corresponding geographical coordinates across the target area.
    \item \textbf{Geographical Space Gridization}: Partition samples into $K$ distinct geographical grids based on spatial proximity and compute the average RSRP $\bar{\vec{y}}_{(k)}$ for each grid.
    \item \textbf{CAPS Estimation}: Assume there are at most $L$ spatial angles containing significant channel paths. For each geographical grid, estimate the average CAPS by solving the following sparse coding problem:
    \begin{equation}\label{eq: channel_estimation}
        \begin{aligned}
            \min_{\left\{\vec{x}_{(k)}\right\}_{k=1}^{K}} &~ \sum_{k=1}^{K}\|\bar{\vec{y}}_{(k)} - \mat{A}\vec{x}_{(k)}\|_2^2,\\
            \mathrm{s.t.} &~ \|\vec{x}_{(k)}\|_0 \le L, \vec{x}_{(k)} \succeq \vec{0}, \, \forall k.
        \end{aligned}
    \end{equation}
    To solve this problem, \cite{zhang2024lscm} developed a weighted nonnegative orthogonal matching pursuit (OMP) algorithm, an enhanced variant of the classical OMP.
\end{enumerate}
Nevertheless, this approach faces two significant practical limitations: (1) the considerable financial cost and time expenditure required for DT data collection, and (2) the limited coverage, as DTs predominantly capture outdoor environments, particularly along roadways, thereby leaving large portions of the BS coverage area unsampled and uncharacterized.

Recent studies \cite{ying2022space} have attempted to address these data limitations by clustering measurement report (MR) samples into grids using beam-level RSRP vectors as the exclusive clustering features, eliminating geographical dependencies. This method is referred to as beam space gridization (BSG), as the beam-level RSRP vector space is termed as \textit{beam space}. However, distinct CAPS patterns can produce similar RSRP measurements \cite{ning2022multi}, which can lead to beam space grids with similar RSRP vectors but different channel characteristics, undermining the fundamental goal of identifying regions with consistent channel behavior.

\section{Problem Formulation}\label{sec:problem_formulation}

In this section, we first introduce the mathematical modeling and assumptions for \emph{Channel Space Gridization} (CSG). Then, we formulate the CSG problem as a joint CAPS estimation and clustering problem.

\subsection{Mathematical Modeling and Assumptions} \label{subsec:CSG_modeling}

Consider $I$ RSRP samples $\set{Y_{I}} \triangleq \left\{\vec{y}_i\right\}_{i=1}^{I}$ collected from MR data under fixed BS-side antenna parameters. Denote the set of sample indices as $\set{I} \triangleq \{i\}_{i=1}^{I}$. Each sample is associated with an unknown CAPS $\vec{x}_i$, such that 
\begin{equation}\label{eq: samples_model}
    \vec{y}_i = \mat{A}\vec{x}_i, \, \forall i \in \set{I}.
\end{equation}
We assume there exist $K$ distinct CAPS patterns, represented as $\set{X_{(K)}} \triangleq \{\vec{x}_{(k)}\}_{k=1}^{K}$, which are determined by the localized environment surrounding the BS and remain relatively stable over time. Each $\vec{x}_{(k)}$ serves as the center of a CAPS grid. Let $\set{K} \triangleq \{k\}_{k=1}^{K}$ denote the set of grid indices, where the number of grids, $K$, is predetermined. Since CAPS represents the channel gain, we impose the constraints:
\begin{equation}\label{eq: nonnegative}
    \vec{x}_i \succeq \vec{0}, \, \vec{x}_{(k)} \succeq \vec{0}, \, \forall i \in \set{I}, \, \forall k \in \set{K}.
\end{equation}
Due to limited number of scatter and reflection points in the environment, we have the following assumption on the sparsity of CAPS grid centers:
\begin{assumption}[Sparsity of CAPS Grid Centers]\label{assum: sparse}
    Each CAPS grid center $\vec{x}_{(k)}$ is $L$-sparse, meaning
    \begin{equation}\label{eq: sparse}
        \|\vec{x}_{(k)}\|_0 \le L, \, \forall k \in \set{K},
    \end{equation}
    where $L < M$. This implies that there exist at most $L$ spatial angles with significant channel paths for each grid center.
\end{assumption}

The primary goal of CSG is to partition the samples into distinct grids, ensuring that each grid contains samples exhibiting similar CAPS patterns. The gridization process assigns each sample to one of the $K$ grids by evaluating the similarity of $\vec{x}_i$ to the grid centers $\set{X_{(K)}}$. We define the gridization process as follows:
\begin{definition}[Channel Space Gridization]\label{def: gridization}
    The $i$-th CAPS sample, $\vec{x}_i$, is assigned to the $k$-th grid if $\vec{x}_{(k)}$ is the closest grid center to $\vec{x}_i$ in terms of the Euclidean distance, \ie, $\vec{x}_{(k)} = \argmin_{\vec{x} \in \set{X_{(K)}}} \|\vec{x}_i - \vec{x}\|_2$. Consequently, the set of sample indices assigned to the $k$-th grid is defined as
    \begin{equation}\label{eq: sample_in_grid_k}
        \set{I}_{(k)} \triangleq \{i \in \set{I} \mid \vec{x}_{(k)} = \argmin_{\vec{x} \in \set{X_{(K)}}} \|\vec{x}_i - \vec{x}\|_2\}.
    \end{equation}
\end{definition}

By assigning samples to the grid with the most similar CAPS pattern, as defined in Definition~\ref{def: gridization}, the ambient space of the BS is effectively partitioned into $K$ regions, each characterized by a distinct CAPS pattern. Let the grid index of the $i$-th sample be denoted as $k_i$, and $\vec{x}_{(k_i)} \triangleq \argmin_{\vec{x} \in \set{X_{(K)}}} \|\vec{x}_i - \vec{x}\|_2$ be the corresponding grid center. Consequently, each sample CAPS $\vec{x}_i$ can be represented as a perturbed version of its corresponding grid center:
\begin{equation}\label{eq: perturbed_sample}
    \vec{x}_i = \vec{x}_{(k_i)} + \vec{e}_{i}, \, \forall i \in \set{I},
\end{equation}
where $\vec{e}_{i}$ denotes the perturbation vector associated with the $i$-th sample. Let $\set{N}_{(k)}$ denote the support of $\vec{x}_{(k)}$, which includes the indices of its nonzero entries:
\begin{equation}\label{eq: support}
    \set{N}_{(k)} \triangleq \{ n \in \{1, \dots, N\} \mid [\vec{x}_{(k)}]_{n} \neq 0 \}.
\end{equation}
The complement of $\set{N}_{(k)}$, denoted as $\set{N}_{(k)}^{c}$, contains the indices of zero entries in $\vec{x}_{(k)}$. Define the orthogonal projection of a vector $\vec{v}$ onto $\set{N}$ as
\begin{equation}\label{eq: projection}
    [\proj_{\set{N}}(\vec{v})]_{n} \triangleq 
    \begin{cases}
        [\vec{v}]_{n}   & \text{if } n \in \set{N}, \\
        0                & \text{otherwise}.
    \end{cases}
\end{equation}

The perturbation vector $\vec{e}_{i}$ can be decomposed into two components: one aligned with the support of $\vec{x}_{(k_i)}$ and the other orthogonal to it:
\begin{equation}\label{eq: perturbation_decomposition}
    \vec{e}_{i} = \underbrace{\proj_{\set{N}_{(k_i)}}(\vec{e}_{i})}_{\text{dominant path perturbation}} + \underbrace{\proj_{\set{N}_{(k_i)}^{c}}(\vec{e}_{i})}_{\text{non-dominant path perturbation}}.
\end{equation}
Here, $\proj_{\set{N}_{(k_i)}}(\vec{e}_{i})$ captures variations in channel gain along the dominant path directions defined by the support of $\vec{x}_{(k_i)}$, while $\proj_{\set{N}_{(k_i)}^{c}}(\vec{e}_{i})$ represents variations in non-dominant directions, typically arising from background noise, secondary paths caused by temporary scatterers, or interference. 

Each CAPS grid center $\vec{x}_{(k)}$ is regarded as a long-term stable CAPS pattern dominated by primary channel paths, whereas the perturbation vector $\vec{e}_{i}$ reflects short-term environmental fluctuations, such as temporary obstacles or transient changes in the communication environment. Let the perturbation vectors within the $k$-th grid, $\{\vec{e}_{i} \mid i \in \set{I}_{(k)}\}$, be \iid~samples from a random vector $\rvec{e}_{(k)}$, we make the following assumption regarding its distribution:
\begin{assumption}[Zero-Mean Perturbations on Dominant Paths]\label{assum: zero-mean-perturbation}
    The entries of $\rvec{e}_{(k)}$ on the support of $\vec{x}_{(k)}$ have zero mean, \ie,
    \begin{equation}
        \mathbb{E}\left[ \proj_{\set{N}_{(k)}}\left(\rvec{e}_{(k)}\right)\right] = \vec{0}.
    \end{equation}
    Additionally, to ensure the nonnegativity of the CAPS, we also have $\rvec{e}_{(k)} \succeq -\vec{x}_{(k)}$.
\end{assumption}

This assumption is reasonable, as the dominant paths are expected to remain stable over time. By enforcing the zero-mean assumption on the dominant path perturbations, the grid centers $\vec{x}_{(k)}$ are designed to capture the long-term stable characteristics of the dominant channel paths. In contrast, the non-dominant paths, influenced by short-term environmental changes or noise, may exhibit random variations that are inherently unpredictable. In this work, we do not impose any specific conditions on the non-dominant path perturbations.

\begin{figure}
    \centering
    \includegraphics[width=\linewidth]{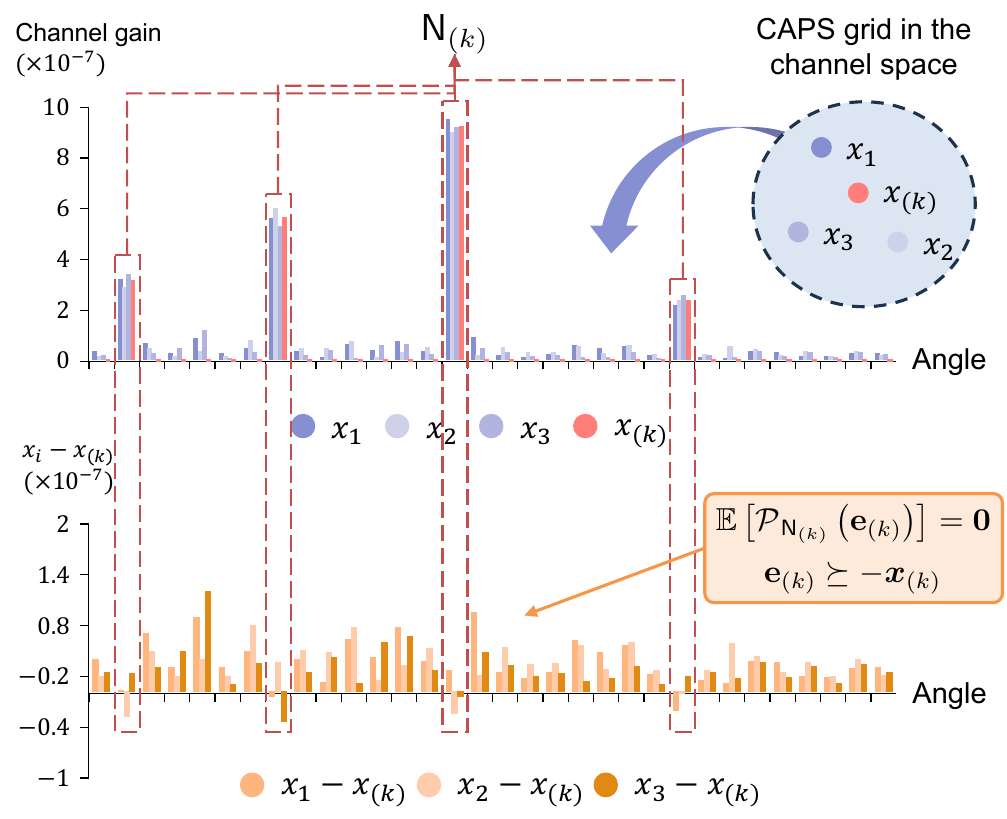}
    \vspace{-1em}
    \caption{An illustration of the zero-mean perturbation assumption.}
    \label{fig:zero-mean-perturbation}
    \vspace{-1em}
\end{figure}

An illustration of the zero-mean perturbation assumption is shown in Fig.~\ref{fig:zero-mean-perturbation}. According to Assumption \ref{assum: zero-mean-perturbation}, with a sufficiently large number of samples within each CAPS grid, we can make the following approximation:
\begin{equation}\label{eq: sample_grid_center}
    \frac{1}{|\set{I}_{(k)}|}\sum_{i \in \set{I}_{(k)}} \proj_{\set{N}_{(k)}} \left(\vec{x}_i - \vec{x}_{(k)}\right) \approx \vec{0}, \, \forall k \in \set{K},
\end{equation}
which implies that the average perturbation of the CAPS samples within each grid is approximately zero on the support of the corresponding grid center. Furthermore, we define a projected average center of the $k$-th grid as
\begin{equation}
    \vec{\mu}_{(k)} \triangleq \frac{1}{|\set{I}_{(k)}|}\sum_{i \in \set{I}_{(k)}} \proj_{\set{N}_{(k)}} \left(\vec{x}_i\right),
\end{equation}
then \eqref{eq: sample_grid_center} can be rewritten as $\vec{\mu}_{(k)} \approx \vec{x}_{(k)}$.

\subsection{Optimization Problem Formulation}\label{subsec:CSG_problem}

The primary objective of CSG is to partition the RSRP samples into CAPS grids. However, achieving this requires knowledge of both the CAPS grid centers $\set{X_{(K)}}$ and the CAPS samples $\set{X_{I}}$, which are unknown in practice. To address this, we aim to jointly estimate these two unknowns while ensuring the properties outlined in subsection \ref{subsec:CSG_modeling} are satisfied. This leads to the following informal formulation of the CSG problem:
\begin{align}
    \mathrm{find} 
    &~~ \set{X_{(K)}} \subset \R^{N} ,\, \set{X_{I}} \subset \R^{N}, \notag \\
    \mathrm{s.t.}
    &~~ 
    \vec{y}_i = \mat{A}\vec{x}_i, \, \, \forall i \in \set{I}, \label{eq: CSG_problem_informal_data_fidelity} \tag{c1}\\
    &~~ 
    \vec{x}_{(k)} \succeq \vec{0}, \, \forall k \in \set{K}, \, \vec{x}_{i} \succeq \vec{0}, \, \forall i \in \set{I}, \label{eq: CSG_problem_informal_nonnegativity} \tag{c2}\\
    &~~ 
    \|\vec{x}_{(k)}\|_0 \le L, \, \forall k \in \set{K}, \label{eq: CSG_problem_informal_sparsity} \tag{c3}\\
    &~~ 
    \vec{\mu}_{(k)} \approx \vec{x}_{(k)}. \label{eq: CSG_problem_informal_zero_mean_approx} \tag{c4}
\end{align}
The constraints serve the following purposes: \eqref{eq: CSG_problem_informal_data_fidelity} ensures accurate RSRP reconstruction from CAPS samples via the LSCM as in \eqref{eq: samples_model}; \eqref{eq: CSG_problem_informal_nonnegativity} enforces nonnegativity of CAPS grid centers and samples as in \eqref{eq: nonnegative}; \eqref{eq: CSG_problem_informal_sparsity} imposes sparsity on grid centers as in \eqref{eq: sparse}; and \eqref{eq: CSG_problem_informal_zero_mean_approx} ensures that the average perturbation of the CAPS samples within each grid is approximately zero on the support of the corresponding CAPS grid center, in line with \eqref{eq: sample_grid_center} derived from Assumption \ref{assum: zero-mean-perturbation}.

To formulate the CSG problem as a well-posed optimization problem, we define appropriate loss functions and incorporate the necessary constraints. In real-world scenarios, RSRP samples are typically collected in dBm units, while the LSCM operates in the linear scale, where RSRP samples are represented in mW units, \ie, $\vec{y}_i^{\mathrm{mW}} = \mat{A}\vec{x}_{i}$. The industry standard for evaluating RSRP estimation accuracy is the mean absolute error (MAE) in the dB scale, defined as $\mathrm{MAE}\left(\vec{y}_1^{\mathrm{dBm}}, \vec{y}_2^{\mathrm{dBm}}\right) \triangleq \frac{1}{M} \left\| \vec{y}_1^{\mathrm{dBm}} - \vec{y}_2^{\mathrm{dBm}} \right\|_1.$
To ensure accurate RSRP estimation, we define a data fidelity loss function based on the MAE between the observed and predicted RSRP samples, which is given by:
\begin{equation}\label{eq: CSG_problem_data_fidelity_loss}
    \Ls_{1}(\set{X_{I}}; \mat{A}, \set{Y_{I}}) = \frac{1}{|\set{I}|} \sum_{i\in\set{I}} \mathrm{MAE}\left(\vec{y}_i^{\mathrm{dBm}}, 10 \log_{10} \left(\mat{A}\vec{x}_{i} \right)\right).
\end{equation}

Precise CAPS estimation is critical, as small fluctuations in CAPS can lead to significant changes in RSRP. To this end, we employ the Mean Squared Error (MSE) to measure the approximation of grid centers, as it imposes a larger penalty on significant errors. The MSE between CAPS is defined as $\mathrm{MSE}\left(\vec{x}_1, \vec{x}_2\right) \triangleq \frac{1}{N} \left\| \vec{x}_1 - \vec{x}_2 \right\|_2^2$. To enforce the zero-mean perturbation on dominant paths, we define the second objective as:
\begin{equation}\label{eq: CSG_problem_zero_mean_approx_loss}
    \Ls_{2}(\set{X_{I}}, \set{X_{(K)}}) = \frac{1}{|\set{K}|} \sum_{k\in\set{K}} \mathrm{MSE} \left( \vec{x}_{(k)}, \vec{\mu}_{(k)}\right).
\end{equation}
This loss quantifies the deviation of the average perturbation of CAPS samples within each grid from the corresponding grid center. Minimizing this loss encourages that the average perturbation is close to zero on the support of the grid center, thereby satisfying the zero-mean assumption.

The objectives \eqref{eq: CSG_problem_data_fidelity_loss} and \eqref{eq: CSG_problem_zero_mean_approx_loss} align with the constraints \eqref{eq: CSG_problem_informal_data_fidelity} and \eqref{eq: CSG_problem_informal_zero_mean_approx}, respectively. Incorporating the remaining constraints \eqref{eq: CSG_problem_informal_nonnegativity} and \eqref{eq: CSG_problem_informal_sparsity}, the CSG problem can be expressed as:
\begin{equation}\label{eq: CSG_problem_formal}\tag{CSG}
    \begin{aligned}
        \min_{\set{X_{(K)}} \subset \R^{N}, \set{X_{I}} \subset \R^{N}} &~~ w_1 \Ls_{1}(\set{X_{I}}; \mat{A}, \set{Y_{I}}) + w_2 \Ls_{2}(\set{X_{I}}, \set{X_{(K)}}), \\
        \mathrm{s.t.~~~~~~}
        &~~
        \vec{x}_{(k)} \succeq \vec{0}, \, \forall k \in \set{K}, \, \vec{x}_{i} \succeq \vec{0}, \, \forall i \in \set{I}, \\
        &~~ 
        \|\vec{x}_{(k)}\|_0 \le L, \, \forall k \in \set{K},
    \end{aligned}
\end{equation}
where $w_1$ and $w_2$ are non-negative weights balancing the relative importance of each objective. The formulation \eqref{eq: CSG_problem_formal} provides a unified framework to jointly determine the CAPS grid centers $\set{X_{(K)}}$ and CAPS samples $\set{X_{I}}$.

The minimization of $\Ls_{2}$ can be framed as a clustering problem, where $\set{I}_{(1)}, \dots, \set{I}_{(K)}$ denotes the assignment of samples to clusters. Solving problem \eqref{eq: CSG_problem_formal} exactly is challenging due to its NP-hard nature, arising from the inherent clustering task compounded by multiple objectives and specific constraints. Notably, there also exists a complex interplay between variables and objective $\Ls_2$: changes in $\set{X_{I}}$ and $\set{X_{(K)}}$ influence $\set{I}_{(1)}, \dots, \set{I}_{(K)}$ and $\set{N}_{(1)}, \dots, \set{N}_{(K)}$, which subsequently impact the objective $\Ls_2$.

\section{Proposed Method}\label{sec:proposed_method}

To tackle this challenge, we propose a deep learning framework, referred to as Channel Space Gridization Autoencoder (CSG-AE), which employs an AE-like architecture to solve the CSG problem effectively.

\begin{figure}[tb]
    \centering
    \includegraphics[width=\linewidth]{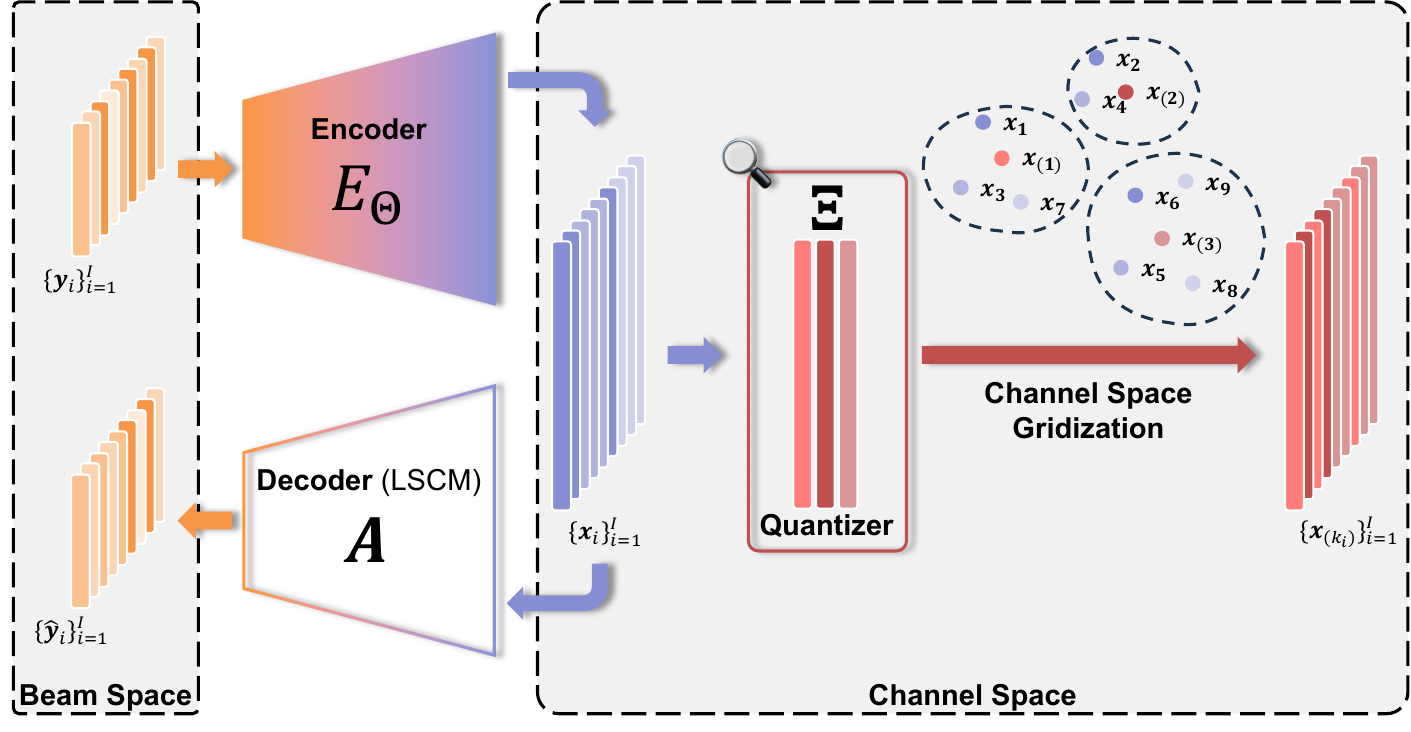}
    \vspace{-1em}
    \caption{The architecture of the CSG-AE. The encoder is a neural network that maps RSRP samples to CAPS samples, while the quantizer maps CAPS samples to grid centers. The decoder is the LSCM that reconstructs RSRP samples from CAPS samples.}
    \label{fig: CSG-AE}
    \vspace{-1em}
\end{figure}

\subsection{Channel Space Gridization Autoencoder}\label{subsec:CSG_AE}

With the benefit of large-scale MR data, we propose a deep learning framework to address \eqref{eq: CSG_problem_formal} in an unsupervised manner. This framework is realized through an AE-like architecture, termed the Channel Space Gridization Autoencoder (CSG-AE). The architecture of the CSG-AE is depicted in Fig.~\ref{fig: CSG-AE}. The CSG-AE consists of three core components:

\paragraph{\textbf{Encoder}}
The encoder, denoted as $E_{\Theta}(\cdot)$, is a neural network (NN) parameterized by $\Theta$, designed to map RSRP samples to CAPS samples. To ensure the non-negativity constraint \eqref{eq: CSG_problem_informal_nonnegativity} is satisfied, a ReLU activation function is applied to the NN's output. The estimated CAPS for the $i$-th sample is thus defined as:
\begin{equation}
    \vec{x}_i \triangleq E_{\Theta}(\vec{y}_{i}^{\mathrm{dBm}}) = \ReLU\left(g_{\Theta}(\vec{y}_i^{\mathrm{dBm}})\right),
\end{equation}
where $g_{\Theta}(\cdot): \R^{M} \rightarrow \R^{N}$ is the NN within the encoder.

\paragraph{\textbf{Quantizer}}
The quantizer, $Q_{\mat{\Xi}}(\cdot)$, maps the encoder's output to the nearest CAPS grid center. It is parameterized by a learnable codebook $\mat{\Xi} \in \R^{N \times K}$, where each column $\vec{\xi}_{k}$ represents a codeword. Let $\mat{\Xi}[k]$ denote the operation of querying the $k$-th codeword from the codebook. To satisfy the constraints \eqref{eq: CSG_problem_informal_nonnegativity} and \eqref{eq: CSG_problem_informal_sparsity}, each queried codeword undergoes $L$-sparse hard thresholding followed by ReLU activation, \ie, $\mat{\Xi}[k] \triangleq \ReLU\left(\delta_{L}\left( \vec{\xi}_{k} \right)\right)$, where $\delta_{L}(\cdot)$ retains the $L$ largest elements of the input vector and sets the remaining elements to zero. The quantizer identifies the nearest codebook entry to the encoder's output, with the grid index of the $i$-th sample determined as $k_i \triangleq \arg\min_{k \in \set{K}} \left\|\vec{x}_i - \mat{\Xi}[k]\right\|_2$. Consequently, the estimated CAPS grid center for the $i$-th sample is:
\begin{equation}\label{eq: grid_center_i_sample}
    \vec{x}_{(k_i)} \triangleq Q_{\mat{\Xi}}(\vec{x}_i) = \mat{\Xi}[k_i],
\end{equation}
and the set of sample indices assigned to the $k$-th grid is:
\begin{equation}\label{eq: grid_index_i_sample_set}
    \set{I}_{(k)} \triangleq \{i \in \set{I} \mid \mat{\Xi}[k] = Q_{\mat{\Xi}}\left(\vec{x}_i\right)\}.
\end{equation}

\paragraph{\textbf{Decoder}}
The decoder, $D(\cdot)$, leverages the LSCM~\eqref{eq: LSCM}, a deterministic mapping from CAPS to RSRP, to reconstruct RSRP samples from encoder's outputs:
\begin{equation}
    \hat{\vec{y}}_i^{\mathrm{dBm}} \triangleq D(\vec{x}_{i}) = 10 \log_{10} \left(\mat{A}\vec{x}_i \right).
\end{equation}
With the physical model, the decoder ensures that the reconstruction process adheres to the intrinsic channel properties.

By constructing the model described above, we embed the constraints of the CSG problem into the specialized architectures of the CSG-AE, which inherently enforce these constraints during training. Specifically, the encoder and quantizer are meticulously designed to learn a mapping from RSRP samples to CAPS samples and grid centers, respectively. The motivation behind this design is further detailed in Section \ref{subsec:CSG_AE_motivation}. Therefore, the CSG problem \eqref{eq: CSG_problem_formal} can be reformulated as a training problem of the CSG-AE, which minimizes the following loss function:
\begin{equation}\label{eq: CSG_AE_loss}
    \resizebox{0.89\linewidth}{!}{$\displaystyle
    \begin{aligned}
        & \Ls_{\text{CSG-AE}}(\Theta, \mat{\Xi}; \mat{A}, \set{Y_{I}}) \triangleq w_1 \Ls_{1}(\Theta; \mat{A}, \set{Y_{I}}) + w_2 \Ls_{2}(\Theta, \mat{\Xi}; \set{Y_{I}}) \\
        & = \frac{w_1}{|\set{I}|} \sum_{i\in\set{I}}  \mathrm{MAE} \left( \vec{y}_i^{\mathrm{dBm}}, \hat{\vec{y}}_i^{\mathrm{dBm}} \right) + \frac{w_2}{|\set{K}|} \sum_{k \in \set{K}} \mathrm{MSE}\left( \vec{x}_{(k)}, \vec{\mu}_{(k)} \right),
    \end{aligned}
    $}
\end{equation}
where we have $\hat{\vec{y}}_i^{\mathrm{dBm}} \triangleq 10 \log_{10} \left(\mat{A}E_{\Theta}(\vec{y}_{i}^{\mathrm{dBm}})  \right)$, $\vec{\mu}_{(k)} \triangleq \frac{1}{|\set{I}_{(k)}|}\sum_{i \in \set{I}_{(k)}} \proj_{\set{N}_{(k)}} \left(E_{\Theta}(\vec{y}_i^{\mathrm{dBm}})\right)$ with $\set{I}_{(k)} \triangleq \{i \in \set{I} \mid \mat{\Xi}[k] = Q_{\mat{\Xi}}\left(E_{\Theta}\left(\vec{y}_i^{\mathrm{dBm}}\right)\right)\}$. 
The loss function consists of two terms, each addressing a distinct aspect of the CSG problem:
\begin{itemize}[leftmargin=*]
    \item \textbf{Reconstruction Loss} ($\Ls_1$): This term measures how well the encoder's output reconstructs the original RSRP samples through the LSCM decoder. It ensures that the learned CAPSs are physically meaningful under the channel model and consistent with the observed data.
    \item \textbf{Quantization Loss} ($\Ls_2$): This term measures the discrepancy between each grid center and the average of the projected encoder's outputs assigned to it. It promotes grid centers to be representative of their associated samples.
\end{itemize}

The CSG-AE architecture supports end-to-end training, enabling joint optimization of the encoder, quantizer, and decoder parameters by minimizing the loss function \eqref{eq: CSG_AE_loss} via backpropagation. However, to overcome specific training challenges in naive training scheme, we propose an enhanced training scheme, detailed in Section \ref{subsec:training_scheme}.

\subsection{Motivation for Applying Learning-Based Method}\label{subsec:CSG_AE_motivation}

The CSG problem \eqref{eq: CSG_problem_formal} can be addressed using various optimization techniques, such as alternating minimization \cite{bertsekas1999nonlinear, yang2019inexact} and proximal gradient descent \cite{parikh2014proximal}. However, directly estimating the unknown CAPS variables and grid centers using these traditional approaches is computationally intensive, particularly when the number of samples is large. Moreover, as new data is continuously generated, these methods require re-solving the optimization problem for each new dataset, rendering them impractical in real-world scenarios.

In contrast, we propose to learn the mapping from RSRP samples to their corresponding CAPS, thereby eliminating the need to explicitly optimize a large set of sample-specific variables and to repeatedly optimize for each new data instance. Specifically, we posit two mapping functions: one from RSRP samples to CAPS samples, denoted as $h_1(\cdot)$, and another from CAPS samples to grid centers, denoted as $h_2(\cdot)$, such that $\vec{x}_{i} = h_1(\vec{y}_i)$ and $\vec{x}_{(k_i)} = h_2(\vec{x}_{i})$, $\forall i \in \set{I}$. These mapping functions must adhere to the constraints \eqref{eq: CSG_problem_informal_nonnegativity} and \eqref{eq: CSG_problem_informal_sparsity}. In our design, we realize $h_1(\cdot)$ with an encoder network $h_1(\cdot) \approx E_{\Theta}(\cdot)$ and $h_2(\cdot)$ with a quantizer network \cite{van2017vqvae} $h_2(\cdot) \approx Q_{\mat{\Xi}}(\cdot)$. Both networks are architected so that their outputs automatically satisfy the required constraints. The CSG problem is then reformulated as a learning task:
\begin{equation}\label{eq: CSG_problem_formal_amortized}\tag{CSG-AE}
        \min_{\Theta, \mat{\Xi}} ~~ \Ls_{\text{CSG-AE}}(\Theta, \mat{\Xi}; \mat{A}, \set{Y_{I}}),
\end{equation}
where we express the unknown sample-specific variables $\set{X_{I}}$ and $\set{X_{(K)}}$ through shared parameterized models with a fixed number of learnable parameters $\Theta$ and $\mat{\Xi}$. Consequently, no explicit per-sample optimization is required. Instead, large-scale instances can be processed in parallel on GPU/TPU, which significantly improves computational efficiency.

Once the model is trained on a large dataset of RSRP samples, only one forward calculation is needed during deployment, offering a substantial speed advantage over traditional iterative methods. Specifically, after optimizing the parameters $\Theta^{*}, \mat{\Xi}^{*} = \argmin_{\Theta, \mat{\Xi}} \Ls_{\text{CSG-AE}}(\Theta, \mat{\Xi}; \mat{A}, \set{Y_{I}})$, for a new dataset $\set{Y_{I}^{\prime}}$ collected from the same BS under nearly unchanged localized environment, the corresponding CAPS $\set{X_{I}^{\prime}}$ and grid centers $\set{X_{(K)}^{\prime}}$ can be computed efficiently as $\vec{x}_{i}^{\prime} = E_{\Theta^{*}}(\vec{y}_i^{\prime})$ and $\vec{x}_{(k_i)}^{\prime} = Q_{\mat{\Xi}^{*}}(\vec{x}_{i}^{\prime})$, $\forall i \in \set{I}^{\prime}$. Indeed, if environmental conditions change significantly, the model should be retrained on newly acquired data to update its parameters. Nevertheless, owing to the long-term stability of wireless propagation environments, the learned parameters typically remain effective for a long time without frequent retraining.

This follows the principle of \emph{amortized optimization} \cite{rui2017amortized, amos2023tutorial}---the heavy computation is paid once to fit a shared inference model, after which each new instance is solved by a single forward pass. From an operational viewpoint, the one-off training cost is amortized over a long time of the network operation, making the proposed learning-based CSG framework far more scalable and cost-effective for large-scale network optimization tasks, which motivates us to apply learning-based method to address the CSG problem.

\subsection{Training Scheme for CSG-AE}\label{subsec:training_scheme}

\begin{figure*}[t]
    \centering
    \includegraphics[width=\linewidth]{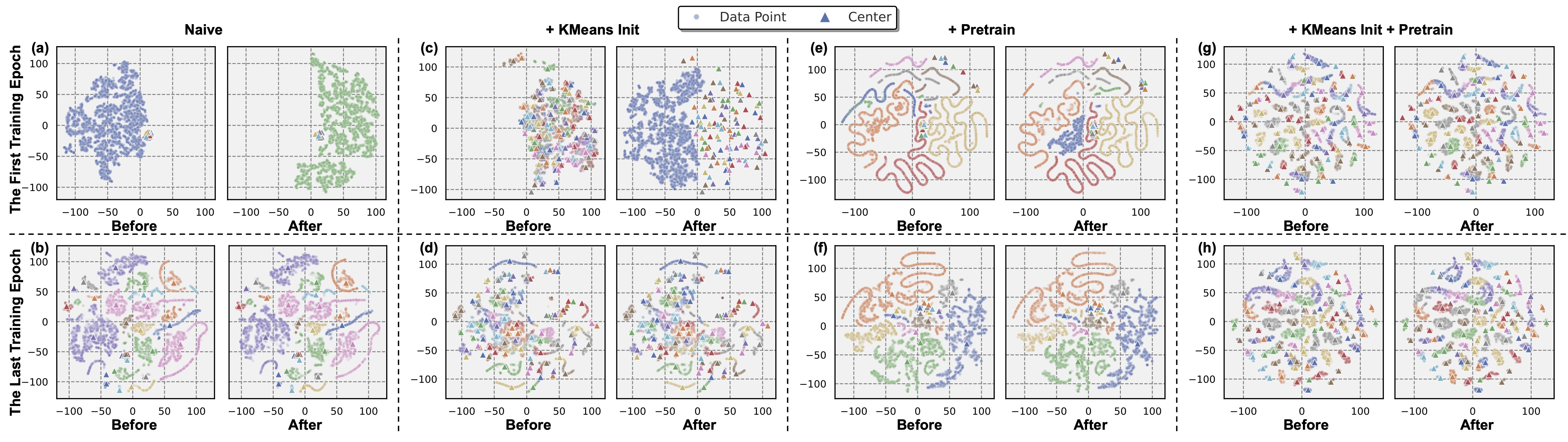}
    \vspace{-1em}
    \caption{Visualization of the centers $\vec{x}_{(k)}$ (triangles) and the projected embeddings $\proj_{\set{N}_{(k_i)}}\left(\vec{x}_{i}\right)$ (circles) using t-SNE~\cite{van2008visualizing}. Colors represent the cluster labels. The first row illustrates the states under different initialization strategies at the beginning of training and immediately after the first training epoch, while the second row depicts the states at the end.}
    \label{fig:tsne-visual}
    \vspace{-1em}
\end{figure*}

\begin{figure}[t]
    \centering
    \includegraphics[width=0.95\linewidth]{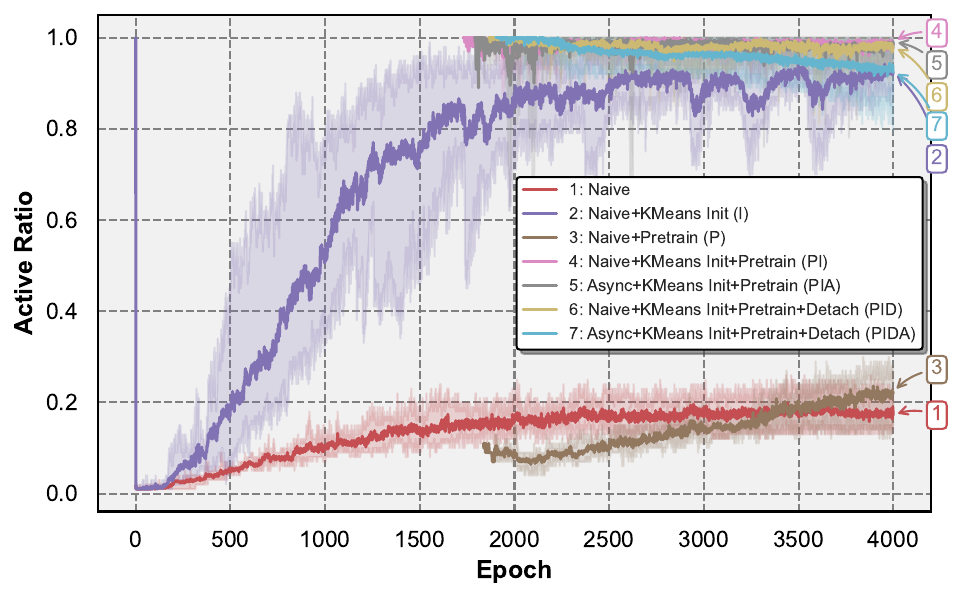}
    \vspace{-1em}
    \caption{Codebook utilization dynamics across training strategies. The naive approach (1) exhibits severe codebook collapse, while our combined $K$-Means initialization and encoder pretraining strategy (4, 5, 6, and 7) maintains near-perfect utilization. For pretraining-based strategies, the encoder is first pretrained for 2000 epochs, with the one achieving the lowest validation loss retained, the full model is then trained from this checkpoint until the end.}
    \label{fig:active-ratio}
    \vspace{-1em}
\end{figure}

As discussed in Section~\ref{subsec:CSG_AE_motivation}, we have reformulated the optimization problem \eqref{eq: CSG_problem_formal} into an end-to-end learning framework, where the model architecture inherently meets the necessary constraints. A straightforward method to training the model involves minimizing the loss function $\Ls_{\text{CSG-AE}}$ via standard backpropagation (see Algorithm \ref{alg: CSG_AE_training_naive} in Appendix \ref{appendix: naive_training_algorithm}). However, this naive strategy often yields suboptimal results due to two key challenges: \emph{low codebook utilization} and \emph{dynamic clustering process}. In this section, we propose an innovative training scheme (detailed in Algorithm \ref{alg: CSG_AE_training}) designed to address and mitigate these challenges effectively.

\subsubsection{Challenges of Naive Training Scheme}
We first briefly discuss these two challenges here. For a detailed discussion of these challenges, please refer to Appendix \ref{appendix: naive_training_algorithm}.

\paragraph{\textbf{Challenge 1---Low Codebook Utilization}}
The naive training approach suffers from \textit{codebook collapse} \cite{kaiser2018fast, roy2018theory, huh2023straightening}, where only a small subset of codewords remain active, as demonstrated by curve 1 in Fig.~\ref{fig:active-ratio}. This phenomenon arises from misalignment between encoder-generated embeddings and codewords, driven by \emph{initialization misalignment}---when codebook vectors are randomly initialized to be far from the encoder embeddings, certain codewords become inactive (see Fig.~\ref{fig:tsne-visual}~(a) and (e), only a few colors are displayed in the data points, indicating that there are only a few active codewords) and their loss terms reduce to MSE with the zero vector, pushing them toward the origin---and \emph{embedding drift}---significant encoder parameter updates in early training induce embedding shifts (see Fig.~\ref{fig:tsne-visual}~(a) and (c), data points drift a lot after the first training epoch), disrupting transient codebook-activation patterns before they stabilize.

\paragraph{\textbf{Challenge 2---Dynamic Clustering Process}}
Training CSG-AE involves a bidirectional dynamic clustering process, where both embeddings and grid centers are updated. Temporally, the assignment in the naive training scheme is hysteretic, where grid center updates depend on current embeddings, causing centers to pursue outdated targets. Spatially, gradients with respect to embeddings $\nabla_{\vec{x}_i^{(t)}} \Ls_{1}$ and $\nabla_{\vec{x}_i^{(t)}} \Ls_{2}$ conflict, when the angle between them exceeds $90^\circ$, leading to partial cancellation. While such gradient direction conflicts are not problematic in isolation, their coupling with assignment update hysteresis introduces complex optimization dynamics.

\subsubsection{Solution for Challenge 1---Encoder Pretraining and \texorpdfstring{\kmeans~Codebook Initialization}{K-Means Codebook Initialization}}
To overcome the low codebook utilization issue, we introduce a two-step strategy:
\paragraph{\textbf{Step 1---Encoder Pretraining}}
We first pretrain the encoder for several epochs on the same training dataset, optimizing only the reconstruction loss $\Ls_{1}$. This step enables the encoder to learn preliminary representations of the input data, thereby stabilizing the embedding manifold before the introduction of quantization effects. As shown in Fig.~\ref{fig:tsne-visual}~(e) and (g), the pretrained embeddings exhibit spatial consistency, effectively preventing drift-induced deactivation of codewords.
\paragraph{\textbf{Step 2---\kmeans~Codebook Initialization}}
Following encoder pretraining, we apply the \kmeans~clustering algorithm \cite{lloyd1982least} to the encoder's embeddings to obtain $K$ cluster centroids, which serve as the initial codewords for the quantizer. This initialization ensures that the codewords are well-distributed across the embedding manifold, as illustrated in Fig.~\ref{fig:tsne-visual}~(c) and (g), ensuring an initial active ratio of 100\%. Such geometric alignment provides stable gradient signals for all codewords from the beginning of training.

Ablation studies (see Fig.~\ref{fig:active-ratio}) reveal that neither step alone suffices to fully resolve the misalignment issue. Only their combined application achieves nearly 100\% codebook utilization throughout training. 
Specifically, while \kmeans~initialization raises the final codebook activation ratio from 26\% to 86\%, it does not fully avert codebook collapse. As depicted in Fig.~\ref{fig:tsne-visual}~(c), significant encoder updates during early training cause embeddings to drift away from the initially well-aligned codewords, leading to a rapid drop in the active ratio from 100\% to nearly 1\% (see curve 2 in Fig.~\ref{fig:active-ratio}, a rapid drop happens after the first training epoch). As training continues and encoder updates become less pronounced, codewords gradually realign with the embeddings, partially restoring the active ratio. This highlights the need for encoder pretraining as an additional stabilization mechanism. 
However, relying solely on encoder pretraining yields an active ratio of only 18\%, as the randomly initialized codebook remains misaligned with the pretrained embeddings, leaving many codewords inactive (see Fig.~\ref{fig:tsne-visual}~(e)). 
In contrast, the combined approach achieves remarkable success, maintaining codebook utilization above 95\% during training, underscoring the indispensability of both components for optimal performance.

Together, these two steps constitute a robust initialization strategy for the CSG-AE model: encoder pretraining provides favorable initialization for 
the encoder parameters, while \kmeans~codebook initialization ensures the codewords are well-aligned with the embeddings. This coordinated 
design sustains high codebook utilization throughout training. After initialization, the full CSG-AE model is trained to minimize the combined loss 
$\Ls_{\text{CSG-AE}}$.

\subsubsection{Solution for Challenge 2---Detached and Asynchronous Update}

To address the assignment update hysteresis and gradient direction conflicts, we propose a detached and asynchronous update strategy for training the CSG-AE model.

\paragraph{\textbf{Detached Update}}
To resolve gradient direction conflicts between $\Ls_{1}$ and $\Ls_{2}$ in updating embeddings, we employ a detached update strategy where each loss exclusively governs its associated parameters (we denote the optimizer, e.g., Adam \cite{kingma2014adam}, SGD, etc., as \texttt{opt}):
\begin{equation}
    \resizebox{0.89\linewidth}{!}{$\displaystyle
    \Theta^{(t+1)}\!\gets\!\texttt{opt}(\Theta^{(t)}, \nabla_{\Theta^{(t)}} \Ls_{1}), \, \mat{\Xi}^{(t+1)}\!\gets\!\texttt{opt}(\mat{\Xi}^{(t)}, \nabla_{\mat{\Xi}^{(t)}} \Ls_{2}).
    $}
\end{equation}
To achieve this, $\Ls_{2}$ is computed using detached embeddings:
\begin{equation}
    \resizebox{0.89\linewidth}{!}{$\displaystyle
        \Ls_{2}(\mat{\Xi}; \set{\dot{X}_{I}}, \mat{A}, \set{Y_{I}}) = \frac{1}{|\set{K}|} \sum_{k \in \set{K}} \mathrm{MSE}\left( \mat{\Xi}[k], \frac{\sum_{i \in \set{I}_{(k)}} \proj_{\set{N}_{(k)}} \left(\dot{\vec{x}}_i\right)}{|\set{I}_{(k)}|}  \right),
    $}
\end{equation}
where $\dot{\vec{x}}_i\!\gets\!E_{\Theta}(\vec{y}_i).\texttt{detach()}$ represents the detached embedding. The \texttt{detach()}\footnote{We adopt the operator name ``detach'' from PyTorch; the equivalent operation in TensorFlow or JAX is called ``stop gradient''.} operation blocks gradient flow between encoder parameters and quantization loss by enforcing $\nabla_{\Theta}\dot{\vec{x}}_i=\vec{0}$ mathematically, preventing gradients of $\Ls_{2}$ from backpropagating to the encoder \cite{pytorch_detach}. Consequently, $\Ls_{2}$ exclusively updates codebook parameters while embeddings remain unaffected by quantization loss. This separation ensures embeddings optimize solely for reconstruction while codewords independently adapt to represent grid centers.

\paragraph{\textbf{Asynchronous Update}}
To ensure timely and consistent assignment updates, we propose an asynchronous update strategy. The detached update enables independent optimization of encoder and quantizer parameters, enabling a sequential update scheme where embeddings are updated first, followed by grid centers. After updating the encoder and obtaining the latest embeddings $\dot{\vec{x}}_i^{(t+1)}$, assignments to codewords are determined based on these fresh embeddings:
\begin{equation}
    \set{I}_{(k)}^{(t+1)}\!\gets\!\{i \in \set{I} \mid \vec{x}_{(k)}^{(t)} = Q_{\mat{\Xi}^{(t)}}(\dot{\vec{x}}_{i}^{(t+1)})\}, \forall k \in \set{K},
\end{equation}
where $\vec{x}_{(k)}^{(t)}\!\triangleq\!\mat{\Xi}^{(t)}[k]$. These updated assignments facilitate the computation of current projected average centers $\vec{\mu}_{(k)}^{(t+1)}$, ensuring grid center updates utilize the most recent averages: $\nabla_{\vec{x}_{(k)}^{(t)}} \Ls_{2} \propto \vec{x}_{(k)}^{(t)} - \vec{\mu}_{(k)}^{(t+1)}$, rather than the outdated $\vec{\mu}_{(k)}^{(t)}$.

Together, by detaching the gradient flows of reconstruction and quantization losses while updating the encoder and codebook parameters asynchronously, the proposed strategy effectively resolves the assignment update hysteresis and gradient conflict issues inherent in the naive training shceme. Experimental validation on real-world data demonstrates that the combined detached and asynchronous update approach significantly outperforms the naive synchronous training method, as well as individual implementations of either detached or asynchronous updates alone, as evidenced in Fig.~\ref{fig:real_world_results_prediction}.

\subsubsection{Training Scheme for CSG-AE}

Algorithm~\ref{alg: CSG_AE_training} presents the proposed training scheme for the CSG-AE model, termed \textbf{PIDA}. This training scheme for CSG-AE involves the following key strategies: \textbf{P}retraining the encoder to establish preliminary representations of the input data as initial embeddings, \textbf{I}nitializing the codebook through \kmeans~clustering for a well-aligned starting state, and training the full CSG-AE model using a \textbf{D}etached and \textbf{A}synchronous update strategy.

\begin{algorithm}[htb]
    \caption{Proposed Training Scheme for CSG-AE (\textbf{PIDA})} \label{alg: CSG_AE_training}
    \algsetblock[Name]{For}{EndFor}{}{1em}
    \begin{algorithmic}[1]
        \Require RSRP data  \resizebox{!}{0.94\height}{$\set{Y_{I}}$}, beam pattern matrix  \resizebox{!}{0.94\height}{$\mat{A}$}, weights \resizebox{!}{0.94\height}{$w_1$, $w_2$}, encoder warm-up epochs \resizebox{!}{0.94\height}{$T_{0}$}, maximum epochs \resizebox{!}{0.94\height}{$T$}, optimizer \resizebox{!}{0.94\height}{$\texttt{opt}$}, initial encoder parameters \resizebox{!}{0.94\height}{$\Theta^{(0)}$}.
        \Ensure Trained encoder \resizebox{!}{0.94\height}{$E_{\Theta}(\cdot)$} and quantizer \resizebox{!}{0.94\height}{$Q_{\mat{\Xi}}(\cdot)$}.
        \Statex \underline{\Comment{Pretraining Phase} (\textbf{P})}
        \For{ \resizebox{!}{0.94\height}{$t = 0, \dots, T_{0}-1$}}
            \State Encoder:  \resizebox{!}{0.94\height}{$\vec{x}_i^{(t)}\!\gets\!E_{\Theta^{(t)}}(\vec{y}_i), \forall i \in \set{I}$}
            \State Decoder: \resizebox{!}{0.94\height}{$\hat{\vec{y}}_{i}^{(t)}\!\gets\!D(\vec{x}_i^{(t)}), \forall i \in \set{I}$}
            \State Compute reconstruction loss \resizebox{!}{0.94\height}{$\Ls_{1}(\Theta^{(t)}; \mat{A}, \set{Y_{I}})$}
            \State Compute gradients \resizebox{!}{0.94\height}{$\nabla_{\Theta^{(t)}} \Ls_{1}$}
            \State Update \resizebox{!}{0.94\height}{$\Theta^{(t)}$}: \resizebox{!}{0.94\height}{$\Theta^{(t+1)}\!\gets\!\texttt{opt}(\Theta^{(t)}, \nabla_{\Theta^{(t)}} \Ls_{1})$}
        \EndFor
        \Statex \underline{\Comment{Initialization} (\textbf{I})}
        \State Encoder: \resizebox{!}{0.94\height}{$\vec{x}_i^{(T_{0})}\!\gets\!E_{\Theta^{(T_{0})}}(\vec{y}_i), \forall i \in \set{I}$}
        \State Initialize  \resizebox{!}{0.94\height}{$\mat{\Xi}^{(T_{0})}$}: \resizebox{!}{0.94\height}{$\mat{\Xi}^{(T_{0})}\!\gets\!$ \Call{\kmeans}{$\set{X_{I}}^{(T_{0})}, K$}}
        \Statex \underline{\Comment{Training Phase} (\textbf{DA})}
        \For{ \resizebox{!}{0.94\height}{$t = T_{0}, \dots, T-1$}}
            \State Encoder: \resizebox{!}{0.94\height}{$\vec{x}_i^{(t)}\!\gets\!E_{\Theta^{(t)}}(\vec{y}_i), \forall i \in \set{I}$}
            \State Decoder: \resizebox{!}{0.94\height}{$\hat{\vec{y}}_{i}^{(t)}\!\gets\!D(\vec{x}_i^{(t)}), \forall i \in \set{I}$}
            \State Compute reconstruction loss \resizebox{!}{0.94\height}{$\Ls_{1}(\Theta^{(t)}; \mat{A}, \set{Y_{I}})$}
            \State Compute gradients \resizebox{!}{0.94\height}{$\nabla_{\Theta^{(t)}} \Ls_{1}$}
            \State Update \resizebox{!}{0.94\height}{$\Theta^{(t)}$}: \resizebox{!}{0.94\height}{$\Theta^{(t+1)}\!\gets\!\texttt{opt}(\Theta^{(t)}, \nabla_{\Theta^{(t)}} \Ls_{1})$}
            \State Encoder: \resizebox{!}{0.94\height}{$\dot{\vec{x}}_i^{(t+1)}\!\gets\!E_{\Theta^{(t+1)}}(\vec{y}_i)\texttt{.detach()}, \forall i \in \set{I}$}
            \State Quantizer: \resizebox{!}{0.94\height}{$\set{I}_{(k)}^{(t+1)}\!\gets\!\{i \in \set{I} \mid \vec{x}_{(k)}^{(t)}\!=\!Q_{\mat{\Xi}^{(t)}}(\dot{\vec{x}}_{i}^{(t+1)})\}, \forall k \in \set{K}$}
            \State Compute quantization loss \resizebox{!}{0.94\height}{$\Ls_{2}(\mat{\Xi}^{(t)}; \set{\dot{X}_{I}}^{(t+1)}, \mat{A}, \set{Y_{I}})$}
            \State Compute gradients \resizebox{!}{0.94\height}{$\nabla_{\mat{\Xi}^{(t)}} \Ls_{2}$}
            \State Update \resizebox{!}{0.94\height}{$\mat{\Xi}^{(t)}$}: \resizebox{!}{0.94\height}{$\mat{\Xi}^{(t+1)}\!\gets\!\texttt{opt}(\mat{\Xi}^{(t)}, \nabla_{\mat{\Xi}^{(t)}} \Ls_{2})$}
        \EndFor
    \end{algorithmic}
\end{algorithm}

\subsection{One-Shot Learning: Predict RSRP under Other Beam Patterns}
\label{subsec:prediction_other_beam_pattern}

For communication network optimization scenarios where the goal is to identify the beam pattern $\mat{A}$ that maximizes the average RSRP across all grids, subject to some constraints $\mathcal{C}(\mat{A})$ (e.g., power constraints $\sum_{n \in \set{N}} [\mat{A}]_{m, n} \leq P_{\text{max}}$), the optimization problem can be formulated as:
\begin{equation}\label{eq: opt_beam_pattern}
    \max_{\mat{A} \in \mathcal{C}(\mat{A})} \, \sum_{k \in \set{K}} \bar{\vec{y}}_{(k)} \, \Leftrightarrow \, \max_{\mat{A} \in \mathcal{C}(\mat{A})} \, \sum_{k \in \set{K}} \frac{1}{|\set{I}_{(k)}|} \sum_{i \in \set{I}_{(k)}} \vec{y}_{i},
\end{equation}
where $\bar{\vec{y}}_{(k)}$ denotes the average RSRP in the $k$-th grid, and $\set{I}_{(k)} = \{i \in \set{I} \mid k_i = k\}$ is the set of sample indices assigned to grid $k$. Solving this problem requires evaluating the average RSRP for each grid under any candidate beam pattern $\mat{A}$. As it is infeasible to measure RSRP for all possible beam patterns, a reliable predictive channel model is necessary.

The LSCM is especially well-suited for this application \cite{zhang2024lscm}. Specifically, given a set of RSRP samples $\set{Y_{I}}$ collected under a known beam pattern $\mat{A}$, the average RSRP in the $k$-th grid can be computed as:
\begin{equation}
    \bar{\vec{y}}_{(k)} = \frac{1}{|\set{I}_{(k)}|} \sum_{i \in \set{I}_{(k)}} \vec{y}_{i} = \frac{1}{|\set{I}_{(k)}|} \sum_{i \in \set{I}_{(k)}} \mat{A} \vec{x}_{i} = \mat{A} \bar{\vec{x}}_{(k)},
\end{equation}
where $\bar{\vec{x}}_{(k)} \triangleq \frac{1}{|\set{I}_{(k)}|} \sum_{i \in \set{I}_{(k)}} \vec{x}_{i}$ denotes the average CAPS of the $k$-th grid. Since $\vec{x}_{i}$ and $\set{I}_{(k)}$ are independent of the beam pattern, under the assumption that channel characteristics remain relatively stable over time, the average CAPS $\bar{\vec{x}}_{(k)}$ should remain invariant with respect to changes in the beam pattern. Therefore, to predict the average RSRP under a new beam pattern $\mat{A}^{\text{new}}$, one can directly compute $\bar{\vec{y}}_{(k)}^{\text{new}} = \mat{A}^{\text{new}} \bar{\vec{x}}_{(k)}$.

Our CSG-AE model is designed to learn the mapping from RSRP samples to their corresponding CAPS and grid centers. Once trained on RSRP data collected under a specific beam pattern, the model can predict the average RSRP for any other beam patterns without retraining. Specifically, given RSRP samples $\set{Y_I}$ obtained under the original beam pattern $\mat{A}$, we use the trained CSG-AE model with parameters $\Theta^{*}$ and $\mat{\Xi}^{*}$ to predict the average CAPS for the $k$-th grid as follows:
\begin{equation}
    \bar{\vec{x}}_{(k)}^{*} = \frac{1}{|\set{I}_{(k)}^{*}|} \sum_{i \in \set{I}_{(k)}^{*}} \vec{x}_{i}^{*} = \frac{1}{|\set{I}_{(k)}^{*}|} \sum_{i \in \set{I}_{(k)}^{*}} E_{\Theta^{*}}(\vec{y}_{i}),
\end{equation}
where $\set{I}_{(k)}^{*} = \{i \in \set{I} \mid \mat{\Xi}^{*}[k] = Q_{\mat{\Xi}^{*}}(E_{\Theta^{*}}(\vec{y}_{i}))\}$ defines the set of sample indices assigned to the $k$-th grid by the trained quantizer. After training, storing the predicted average CAPS $\bar{\vec{x}}_{(k)}^{*}$ for each grid $k \in \set{K}$ is sufficient. The average RSRP for each grid under any new beam pattern $\mat{A}^{\text{new}}$ can then be efficiently predicted as $\bar{\vec{y}}_{(k)}^{\text{new}^{*}} = \mat{A}^{\text{new}} \bar{\vec{x}}_{(k)}^{*}$. 
Consequently, the optimization problem in \eqref{eq: opt_beam_pattern} can be reduced to a feasible form:
\begin{equation}
    \max_{\mat{A} \in \mathcal{C}(\mat{A})} \, \mat{A} \sum_{k \in \set{K}} \bar{\vec{x}}_{(k)}^{*}.
\end{equation}
This approach enables rapid evaluation of candidate beam patterns without the need for additional costly RSRP measurements, thereby facilitating efficient beam pattern optimization.

\section{Experiments}\label{sec:experimental_results}
In this section, we conduct experiments on both synthetic and real-world datasets to evaluate the effectiveness of our CSG framework and performance of our CSG-AE model.

\subsection{Results on Simulated Dataset}

\subsubsection{Dataset}
The synthetic dataset is constructed as follows. We first generate $|\set{K}|=100$ noise-free, sparse CAPS grid centers $\set{X_{(K)}} = \{\vec{x}_{(k)}\}_{k \in \set{K}}$, each of dimension $N=6552$. For each $\vec{x}_{(k)}$, $L=5$ entries are randomly selected to be nonzero, with values drawn from $|\rv{x}|$, where $x \sim \mathrm{Laplace}(0, \sqrt{p})$ and $p=10^{-5}$. For each grid $k$, $|\set{I}_{(k)}|=100$ perturbed CAPS samples are generated as $\vec{x}_{i} = \vec{x}_{(k)} + \vec{e}_{i}$, where the perturbation vector $\vec{e}_{i}$ is constructed by first sampling $N$ values $z_1, \dots, z_N$ from the truncated normal distribution $\mathcal{TN}(0, 1; -1, 1)$, and then setting $\vec{e}_{i} = [e_1, \dots, e_N]^{\top}$ with $e_n = z_n \cdot s \cdot \min\left(\vec{x}_{(k)}\right)$ if $n \in \set{N}_{(k)}$ and $e_n = |z_n| \cdot s \cdot \min\left(\vec{x}_{(k)}\right)$ if $n \notin \set{N}_{(k)}$. Here, $s$ is a scale factor in $[0.1, 1]$ controlling the perturbation magnitude, and $\min\left(\vec{x}_{(k)}\right)$ denotes the minimum nonzero entry of $\vec{x}_{(k)}$. This construction guarantees that the perturbation vector $\vec{e}_{i}$ satisfies Assumption~\ref{assum: zero-mean-perturbation}. In total, we generate $100 \times 100$ perturbed CAPS samples $\set{X_{I}}$. Next, we apply the beam pattern matrix $\mat{A}$ shown in Fig.~\ref{fig:SRS_matrixA}, which is collected from an actual BS configuration, to generate synthetic RSRP samples as $\vec{y}_{i} = \mat{A}\vec{x}_{i}$ for all $i \in \set{I}$, where $|\set{I}| = 100 \times 100$. The resulting synthetic dataset comprises the CAPS grid centers $\set{X_{(K)}}$, perturbed CAPS samples $\set{X_{I}}$, corresponding RSRP samples $\set{Y_{I}}$, and grid labels $\{k_i\}_{i\in\set{I}}$.

\begin{figure}[t]
    \centering
    \includegraphics[width=\linewidth]{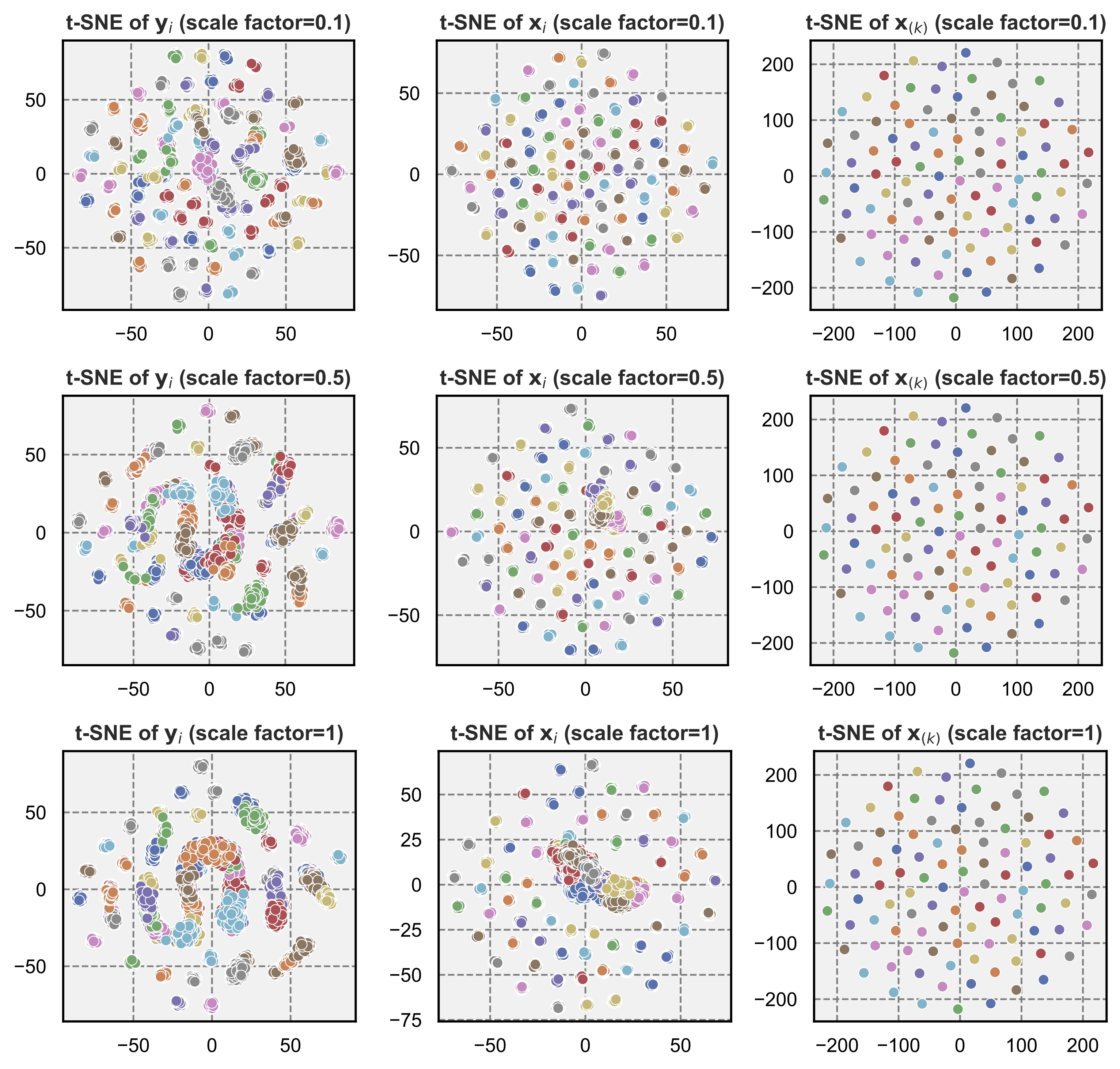}%
    \vspace{-1em}
    \caption{Visualization of synthetic dataset under scale factor $s=0.1, 0.5, 1$ using t-SNE. Points are colored by grid labels.}
    \label{fig:tsne_synthetic_visual}
    \vspace{-1em}
\end{figure}

Fig.~\ref{fig:tsne_synthetic_visual} visualizes some examples from the synthetic dataset with varying scale factors $s$ using t-SNE~\cite{van2008visualizing}. As $s$ increases, the higher perturbation level makes CAPS samples more difficult to cluster. Notably, RSRP samples exhibit poorer clustering properties than CAPS samples, confirming the advantage of performing clustering in the hidden channel space rather than in the observation space.

\subsubsection{Experimental Setup}
The CSG-AE model adopts a 6-layer MLP encoder with 256 hidden units per layer and skip connections at layers 2 and 4. It employs a codebook of size $K=100$ and a linear decoder whose weights are fixed to the beam pattern matrix $\mat{A}$. We employ the proposed PIDA training scheme with AdamW optimizer \cite{loshchilov2017decoupled} using a learning rate of $10^{-2}$ and weight decay of $10^{-4}$. The procedure comprises 2000 epochs of pretraining followed by 2000 epochs of training, with the best models saved at each stage. We compare the proposed CSG-AE with several baselines:
\begin{itemize}[leftmargin=*]
    \item \textsf{KMeans-X/Y}: Apply \kmeans~clustering directly to the perturbed CAPS samples (\textsf{-X}) or RSRP samples (\textsf{-Y}).
    \item \textsf{KMeans-Y-OMP/NOMP/WNOMP}: Apply \kmeans~clustering to RSRP samples, then estimate CAPS centers by applying orthogonal matching pursuit (OMP) or its variants, nonnegative OMP (NOMP) and weighted NOMP (WNOMP) \cite{zhang2024lscm}, to the average RSRP centers of each grid.
    \item \textsf{OMP/NOMP/WNOMP-KMeans-X}: Perform CAPS estimation using OMP, NOMP, or WNOMP, then apply \kmeans~clustering to the estimated CAPS samples.
\end{itemize}
All methods run $5$ times with different random seeds.

\subsubsection{Evaluation Metrics}
We evaluate model performance using two complementary metric categories:
\paragraph{\textbf{Clustering Quality}} With ground-truth grid labels, we employ external clustering metrics: Adjusted Rand Index (ARI), Normalized Mutual Information (NMI), Completeness, Homogeneity, and V-measure. The detailed definitions of these metrics are provided in Appendix \ref{appendix: clustering_metrics}.

\paragraph{\textbf{CAPS Estimation Accuracy}} We measure the CAPS estimation accuracy using the following two metrics:
\begin{gather}
    \text{Sample Mean NMSE} = \frac{1}{|\set{I}|} \sum_{i \in \set{I}} \frac{\|\vec{x}_{(k_i)} - \vec{x}_{(k_i)}^{*}\|_2}{\|\vec{x}_{(k_i)}\|_2}, \\
    \text{Center Wasserstein Distance} = \mathcal{W}\left(\set{X_{(K)}}, \set{X_{(K)}^{*}}\right).
\end{gather}
The Sample Mean NMSE measures the average normalized mean squared error between predicted and ground-truth grid centers, while the Center Wasserstein Distance provides a distribution-level evaluation by measuring the Wasserstein distance $\mathcal{W}$ between predicted and true center distributions.

\subsubsection{Results}

\begin{figure}[tb]
    \centering
    \subfloat[Clustering quality metrics.]{%
        \includegraphics[width=0.98\linewidth]{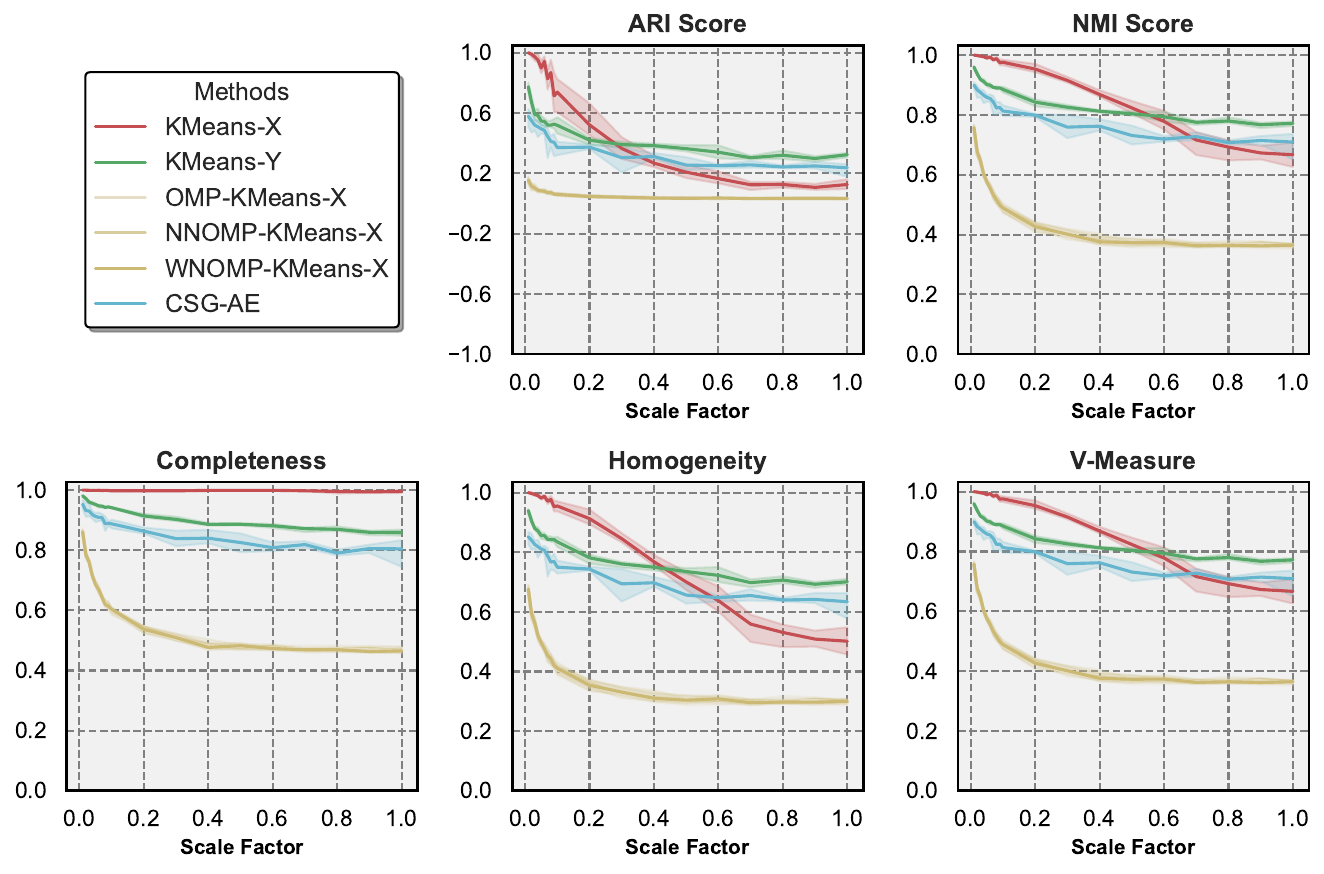}%
        \label{fig:synthetic_results_clustering}
    }
    \vspace{-0.5em}

    \subfloat[CAPS center estimation accuracy metrics.]{%
        \includegraphics[width=0.98\linewidth]{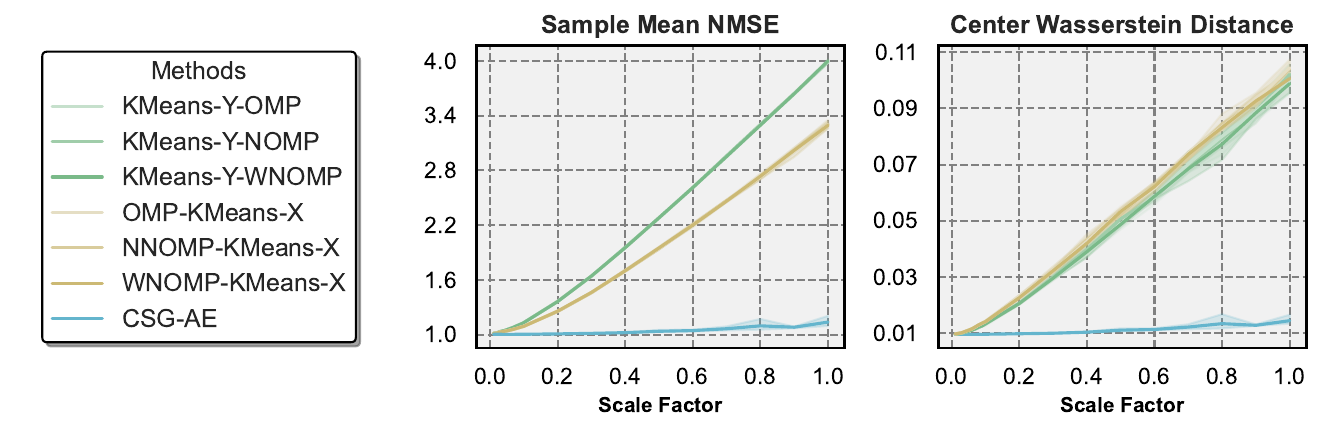}%
        \label{fig:synthetic_results_estimation}
    }
    \caption{Performance comparison of different methods on synthetic dataset. All methods run $5$ times with different random seeds, with the fill-in region in the figures indicating max-min values in $5$ runs.}
    \label{fig:synthetic_results}
    \vspace{-1em}
\end{figure}

Figure~\ref{fig:synthetic_results} presents the evaluation results on the synthetic dataset. The proposed CSG-AE achieves superior CAPS center estimation accuracy and competitive clustering performance compared to all baselines. \textsf{KMeans-X} attains the highest clustering scores for small scale factors $s$, however, its performance deteriorates markedly as $s$ increases. In contrast, both \textsf{KMeans-Y} and CSG-AE maintain robust clustering quality across varying $s$. While \textsf{KMeans-Y} marginally surpasses CSG-AE in clustering metrics, CSG-AE delivers significantly improved CAPS center estimation accuracy at both the sample and distribution levels, underscoring its effectiveness in precise CAPS recovery without sacrificing clustering performance. Sequential methods that estimate CAPS and then cluster (\textsf{OMP/NOMP/WNOMP-KMeans-X}) perform worst in clustering, likely due to error propagation from the initial CAPS estimation step. Unlike these approaches, CSG-AE employs a joint optimization strategy, leveraging neural networks to simultaneously refine CAPS estimation and clustering. This integrated framework enables CSG-AE to achieve superior results in both CAPS center estimation and clustering quality.

\subsection{Results on Real-World Dataset}

\subsubsection{Dataset}
\begin{figure*}[htb]
    \centering
    \includegraphics[width=\linewidth]{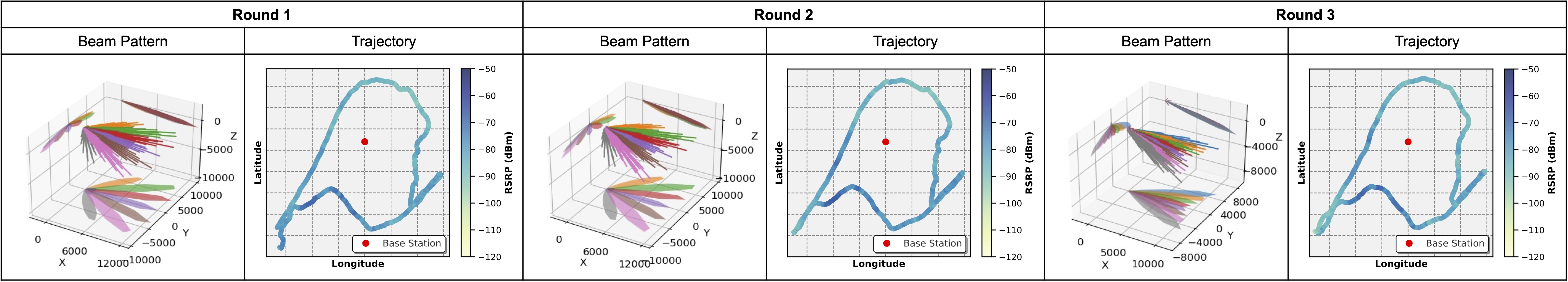}
    \vspace{-1em}
    \caption{Illustration of the walking test dataset collection process (round $1,2,3$). Each round employed a unique beam pattern $\mat{A}^{(r)}, r=1, \dots, 10$, while maintaining the nearly same trajectory, resulting in 10 distinct datasets. The base station is marked by the red point.}
    \label{fig:dataset_detail}
    \vspace{-1em}
\end{figure*}

The real-world dataset was collected via a walking test, where a UE traversed a predefined trajectory around a BS for 10 rounds and the BS applied a unique beam pattern configuration for each round (see Fig.~\ref{fig:dataset_detail}). This procedure produced 10 datasets, for each round $r$, the dataset contains over $10,000$ location measurements $\set{L_{I^{(r)}}}\triangleq\{\vec{l}_{i}\}_{i \in \set{I}^{(r)}}$ and corresponding RSRP measurements $\set{Y_{I^{(r)}}}$, collected under the unique beam pattern $\mat{A}^{(r)}$. We used the first-round dataset for training and reserved the remaining 9 rounds for testing. This allows us to assess the generalization of each method across different beam patterns, reflecting the real-world network optimization scenarios where only data from the current configuration are typically available, as reconfiguration and data recollection are costly and disruptive. Specially, for CSG-AE model training, $10\%$ of the training data was randomly selected as a validation set.

\subsubsection{Experimental Setup}
The CSG-AE model employs the same configuration as in the synthetic dataset experiments, except that the linear decoder's weight matrix is fixed to the first-round beam pattern matrix $\mat{A}^{(1)}$. We employ our proposed PIDA training scheme to optimize the CSG-AE model, complemented by an ablation study examining the impact of detached and asynchronous updates on model performance. We compare the CSG framework with two baselines:
\begin{itemize}[leftmargin=*]
    \item \textbf{GSG}: Following the previous work \cite{zhang2024lscm}, the process of GSG includes: 1) applying \kmeans~clustering to location measurements $\set{L_{I^{(1)}}}$ to form $K=100$ geographical grids; 2) computing the average RSRP for each grid; 3) estimating the average CAPS using OMP and its variants NOMP and WNOMP~\cite{zhang2024lscm} on the average RSRP of each grid.
    \item \textbf{BSG}: This method is applicable when location data are unavailable. Its procedure is identical to GSG, except clustering is performed on RSRP measurements.
\end{itemize}
All methods run $5$ times with different random seeds.

\begin{figure*}[htb]
    \centering

    \subfloat[RSRP prediction error.]{%
        \includegraphics[width=0.47\linewidth]{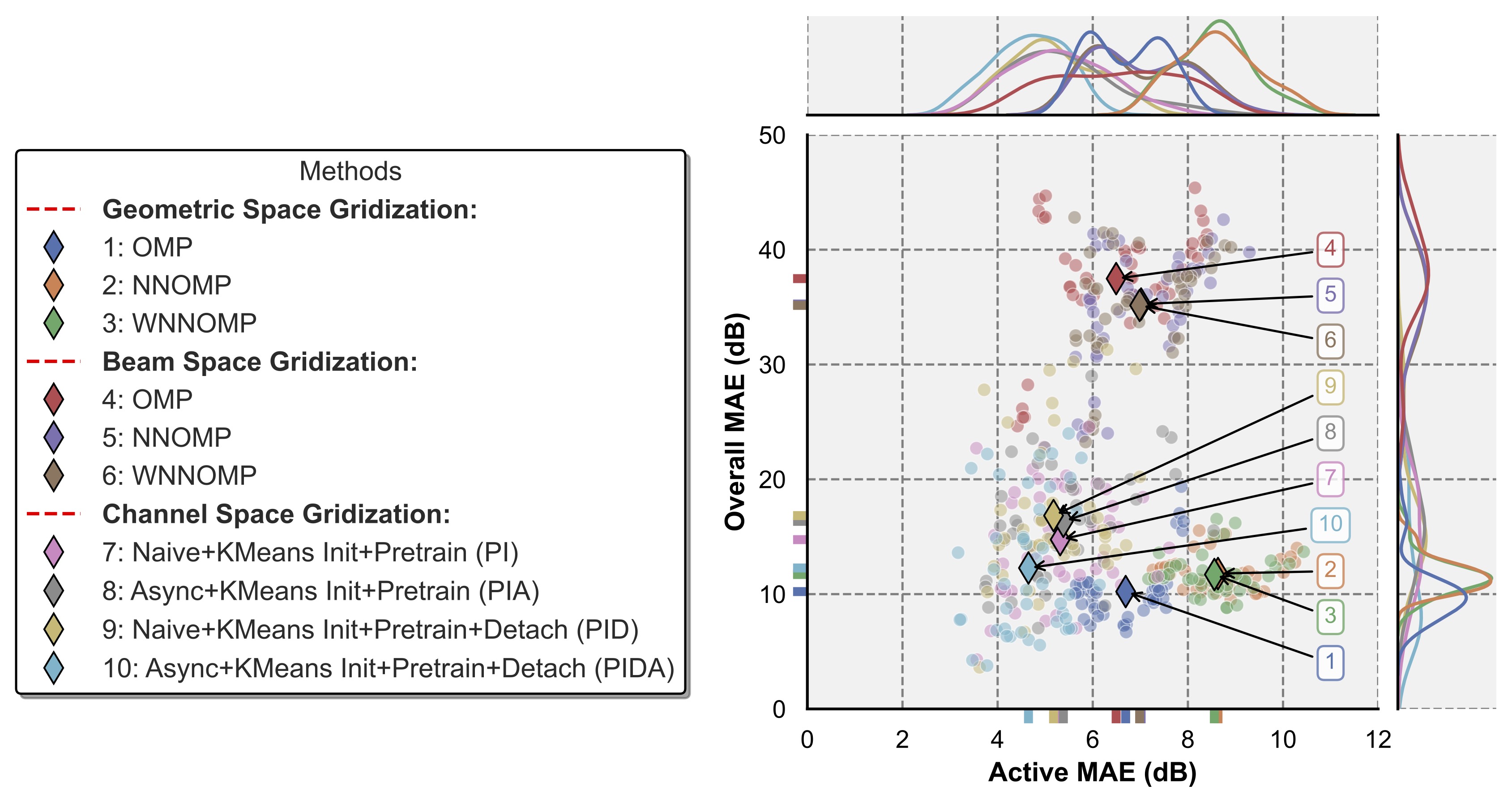}%
        \label{fig:real_world_results_prediction} 
    }
    \vspace{-0.5em}
    \hfill
    \subfloat[Clustering performance.]{%
        \includegraphics[width=0.51\linewidth]{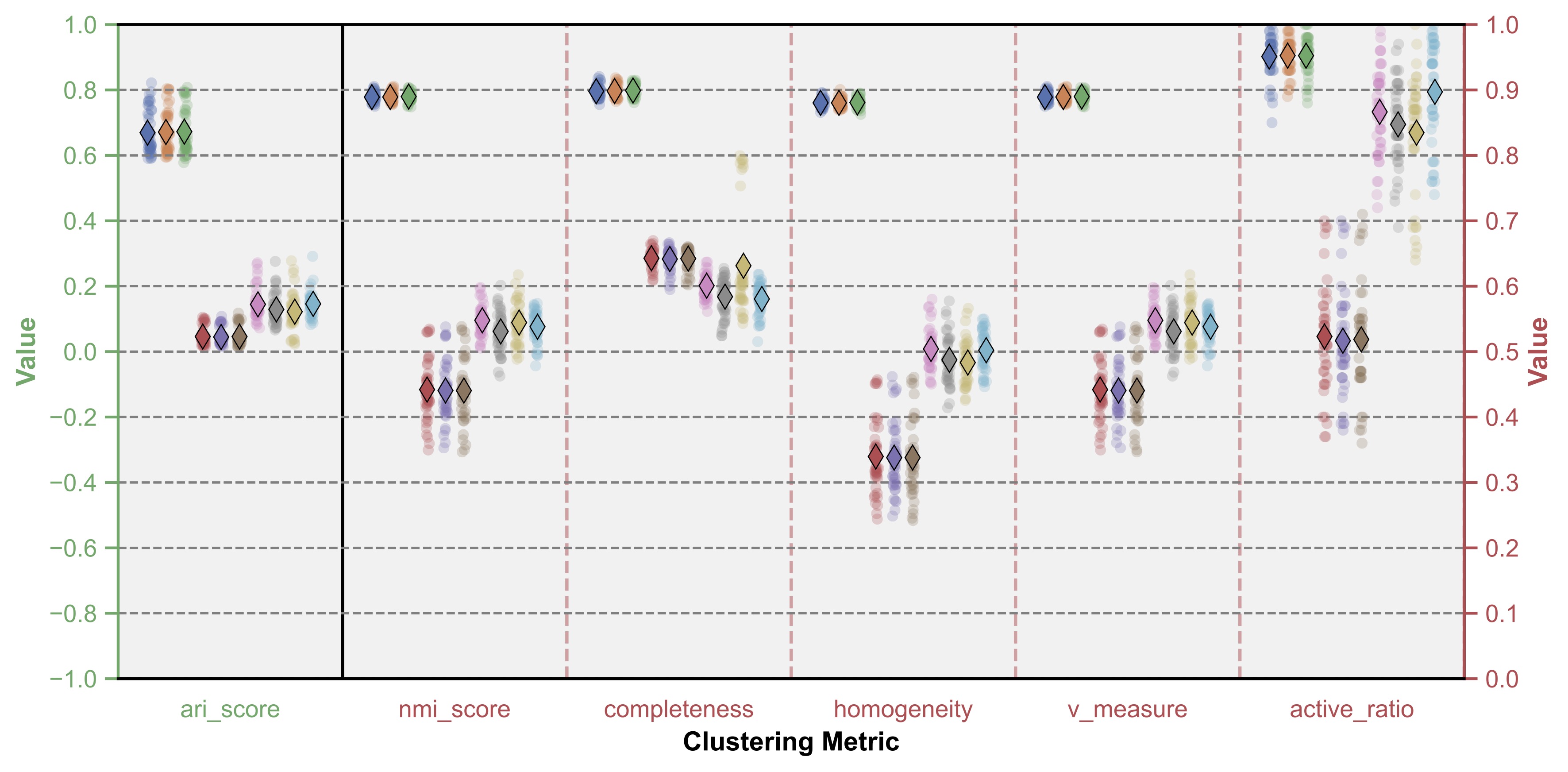}%
        \label{fig:real_world_results_clustering}
    }
    \vspace{-0.5em}

    \subfloat[Distribution of cluster sizes across 9 test rounds.]{%
        \includegraphics[width=0.99\linewidth]{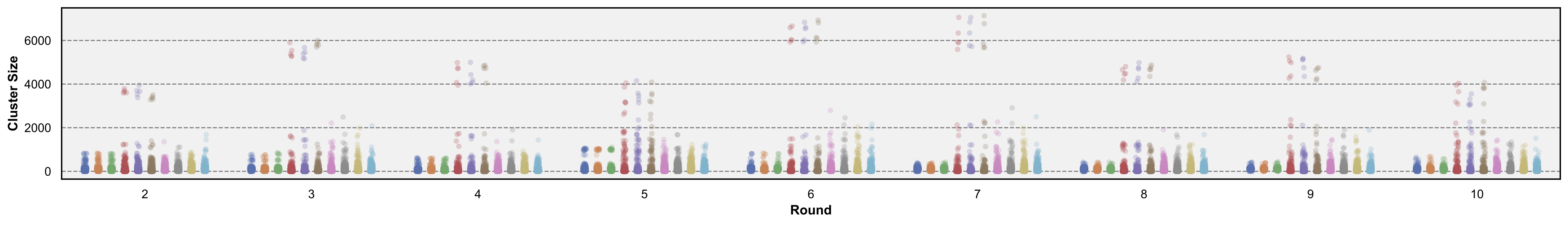}%
        \label{fig:real_world_results_cluster_size}
    }
    \vspace{-0.5em}
    \caption{Performance comparison of different methods on real-world dataset. All methods run $5$ times with different random seeds. In both (a) and (b), circle markers represent performance across test rounds and executions, diamond markers represent the aggregated mean performance, and colors differentiate between methods. In (c), each circle represents a grid's cluster size in a specific round and execution, with colors indicating the corresponding method.}
    \label{fig:real_world_results}
    \vspace{-1em}
\end{figure*}

\subsubsection{Evaluations}
We evaluate the performance of different methods from two key dimensions:

\paragraph{\textbf{RSRP Prediction Accuracy}} 
We quantify CAPS estimation and gridization quality via RSRP prediction accuracy. For each test round, we first predict RSRP values by applying the corresponding beam pattern matrices to estimated CAPS, \ie, $\left(\bar{\vec{y}}_{(k)}^{(r)^{*}}\right)^{\text{dBm}} = 10 \log_{10} \left(\mat{A}^{(r)} \bar{\vec{x}}_{(k)}^{*} \right)$.
We then compute the surrogate ground-truth average RSRP from the real RSRP measurements and the predicted assignments as $\left(\bar{\vec{y}}_{(k)}^{(r)}\right)^{\text{dBm}} = \frac{1}{|\set{I}_{(k)}^{(r)}|} \sum_{i \in \set{I}_{(k)}^{(r)}} \left(\vec{y}_{i}^{(r)}\right)^{\text{dBm}}$, where the grid centers used in $\set{I}_{(k)}^{(r)}$ are those estimated from the first-round dataset. GSG and BSG assign samples by proximity to geographical or beam space grid centers, while CSG uses the trained encoder and quantizer to assign by CAPS proximity. The RSRP prediction accuracy is measured by:
\begin{equation}
    \resizebox{0.89\linewidth}{!}{$\displaystyle
    \begin{aligned}
        &\text{Overall MAE} = \frac{1}{|\set{K}|} \sum_{k \in \set{K}} \text{MAE}\left(\left(\bar{\vec{y}}_{(k)}^{(r)}\right)^{\text{dBm}}, \left(\bar{\vec{y}}_{(k)}^{(r)^{*}}\right)^{\text{dBm}}\right), \\
        &\text{Active MAE} = \frac{1}{|\set{K^{\prime}}|} \sum_{k \in \set{K^{\prime}}} \text{MAE}\left(\left(\bar{\vec{y}}_{(k)}^{(r)}\right)^{\text{dBm}}, \left(\bar{\vec{y}}_{(k)}^{(r)^{*}}\right)^{\text{dBm}}\right),
    \end{aligned}
    $}
\end{equation}
where $\set{K^{\prime}}$ denotes the set of active grids. Active MAE considers only active grids (excluding those with zero real average RSRP), while Overall MAE includes all grids. Lower values of both metrics indicate better CAPS estimation. As grid assignments are method-dependent, these metrics also reflect gridization quality.

\paragraph{\textbf{Clustering Performance}} 
We further evaluate gridization quality by comparing against direct clustering location measurements in term of clustering performance. For each test round, surrogate ground-truth labels are generated by applying \kmeans~to location measurements, serving as proxies for channel similarity---samples with identical labels are expected to exhibit similar channel characteristics. Clustering performance is quantified using external metrics (NMI, ARI, Completeness, Homogeneity, and V-measure). High scores indicate that grid assignments align well with geographical labels, suggesting that grids contain samples with similar channel characteristics. However, the converse is not necessarily true, good gridization performance does not always guarantee high clustering metric scores, thus combining with previous evaluation dimension is necessary for a comprehensive assessment. Additionally, we also report the active ratio, as low values indicate failure to identify and model appropriate grids.

\subsubsection{Results}

Figure~\ref{fig:real_world_results} summarizes the evaluation results on the real-world dataset. GSG, leveraging precise location data, serves as the gold standard. Both BSG and CSG rely solely on RSRP measurements; however, our proposed CSG framework consistently outperforms BSG and, on certain metrics, even exceeds GSG, thereby demonstrating its effectiveness.

\paragraph{\textbf{On RSRP Prediction Accuracy}}
Figure~\ref{fig:real_world_results_prediction} provides a detailed comparison across ten methods in three gridization domains (geographical, beam, and channel) using a joint scatter and distribution plot, with Active MAE on the horizontal axis and Overall MAE on the vertical axis. Notably, CSG-AE with the proposed PIDA training strategy (No. $10$ in Fig.~\ref{fig:real_world_results_prediction}) achieves the best comprehensive performance for both metrics, indicating more accurate CAPS estimation and superior gridization. Specifically,
\begin{itemize}[leftmargin=*]
    \item \textbf{$\mathbf{30\%}$ Improvement in Active MAE.} CSG-AE with PIDA achieves an average Active MAE of approximately $4.6$ dB, outperforming all other methods. This constitutes a $2$ dB ($30\%$) reduction compared to the best BSG (around $6.5$ dB) and GSG (around $6.7$ dB) methods, demonstrating more precise CAPS estimation within active grids.
    \item \textbf{$\mathbf{65\%}$ Improvement in Overall MAE.} CSG-AE with PIDA attains an average Overall MAE of $12.3$ dB, a substantial $22.9$ dB ($65\%$) reduction relative to the best BSG method ($35.2$ dB). Our method's Overall MAE performance is on par with GSG methods, however, it exhibits higher variance, likely attributable to the inherent stochasticity of model training compared to traditional algorithmic approaches. Nevertheless, despite this higher variance, CSG-AE consistently delivers superior performance compared to BSG methods, aligning with our expectation.
\end{itemize}

\paragraph{\textbf{On Clustering Performance}}
Fig.~\ref{fig:real_world_results_clustering} presents a comprehensive comparison of clustering performance across different methods. Despite relying on the same input data, our approach consistently outperforms BSG methods on key clustering metrics. Specifically, it achieves higher ARI values, indicating superior alignment with surrogate labels, and better NMI scores, reflecting more effective information transfer between clusters and labels. More specifically,
\begin{itemize}[leftmargin=*]
    \item \textbf{Higher Within-Grid Channel Similarity.} Although our method yields slightly lower Completeness scores than BSG, it achieves substantially higher Homogeneity and V-measure values. Homogeneity, which quantifies the purity by evaluating whether each cluster contains samples from only one label, is especially relevant for gridization quality in our context. Our method improves Homogeneity by approximately $0.15$ over BSG, indicating that our grids contain more samples sharing the same surrogate label and thus exhibit greater channel similarity.
    \item \textbf{More Balanced Cluster Sizes.} Furthermore, analysis of cluster size distributions (see Fig.~\ref{fig:real_world_results_cluster_size}) shows that BSG methods tend to form larger, imbalanced clusters, explaining their higher Completeness but lower Homogeneity. In contrast, our method produces more balanced cluster sizes, achieving strong Completeness and Homogeneity, leading to better comprehensive gridization performance.
    \item \textbf{Top-Tier Active Ratio.} As shown in Fig.~\ref{fig:real_world_results_clustering}, our method attains an average active ratio of about $90\%$, markedly higher than BSG and close to GSG. This demonstrates that our method has more accurate CAPS estimation thus can maintain a high active ratio across test rounds, consistent with the results in Fig.~\ref{fig:real_world_results_prediction}. The active ratio results also clarifies the discrepancy between BSG's Active MAE and Overall MAE---BSG's low active ratios indicate that many grids are incorrectly classified as inactive, which inflates the Overall MAE and demonstrates their ineffective gridization.
\end{itemize}

\paragraph{\textbf{Unprecedented Gridization Paradigm}} Moreover, as illustrated in Fig.~\ref{fig:real_world_results_clustering}, our method's 
clustering assignments deviate somewhat from the geographic location-based labels as the clustering metrics do not match the performance of 
GSG. This divergence, especially considering our method's superior Active MAE and Overall MAE in Fig.~\ref{fig:real_world_results_prediction},  is intriguing. It implies that our approach may have pioneered a fundamentally different gridization paradigm compared to traditional location-based methods, one that effectively achieves channel-similar gridization without depending on location data.

\section{Conclusion and Future Work}\label{sec:conclusion}
This paper presents CSG, a novel framework that addresses the limitations of traditional gridization methods in wireless network optimization. By integrating channel estimation with gridization, CSG promotes consistent communication characteristics (\ie, CAPS) within each grid. We introduce CSG-AE, a deep learning-based approach that jointly optimizes gridization and channel estimation using readily available RSRP measurements, eliminating the need for expensive DT data or location information. We further propose PIDA, a tailored training scheme that ensures stable training of CSG-AE and overcomes the challenges of naive approaches. Extensive experiments on synthetic and real-world datasets demonstrate CSG-AE's superior performance over BSG, the baseline using identical input data, in both RSRP prediction accuracy and gridization quality. Notably, CSG even surpasses GSG in RSRP prediction, highlighting its effectiveness in achieving channel-consistent gridization bypassing location data.

Looking ahead, we envision several promising directions to advance CSG's capabilities. First, we plan to develop a statistical framework to better address uncertainty and improve generalization, including the use of mixture of Gaussian models \cite{bishop2024discrete} as an alternative to the current \kmeans-based clustering. Second, we aim to extend CSG to multi-cell and multi-frequency scenarios for broader applicability in heterogeneous networks. Third, we will explore leveraging limited location data to further enhance CSG's performance through semi-supervised learning approaches. Fourth, we propose to explore advanced deep learning architectures such as transformer-based models \cite{vaswani2017attention} for capturing temporal variations in channel characteristics and graph neural networks \cite{hamilton2017inductive, yan2023attentional} for modeling spatial correlations between grids. These efforts will position CSG as a foundational framework for environment-aware wireless network intelligence, reducing optimization overhead and improving perception in complex propagation environments.

\appendices


\section{Naive Training Scheme of CSG-AE and Challenges}
\label{appendix: naive_training_algorithm}

The naive training scheme for CSG-AE, as outlined in Algorithm~\ref{alg: CSG_AE_training_naive}, often encounters two significant challenges that impede effective learning and model performance.

\begin{algorithm}[htb]
    \caption{Naive Backpropagation Training of CSG-AE} \label{alg: CSG_AE_training_naive}
    \algsetblock[Name]{For}{EndFor}{}{1em}
    \begin{algorithmic}[1]
        \Require RSRP data $\set{Y_{I}}$, beam pattern matrix $\mat{A}$, weights $w_1$, $w_2$, maximum epochs $T$, optimizer $\texttt{opt}$, initial parameters $\Theta^{(0)}$ and $\mat{\Xi}^{(0)}$.
        \Ensure Trained encoder $E_{\Theta}(\cdot)$ and quantizer $Q_{\mat{\Xi}}(\cdot)$.
        \For{$t = 0, \dots, T-1$}
            \Statex \hspace{0.5em} \underline{\Comment{Forward pass}}
            \State Encoder: $\vec{x}_i^{(t)}\!\gets\!E_{\Theta^{(t)}}(\vec{y}_i), \forall i \in \set{I}$
            \State Quantizer: $\set{I}_{(k)}^{(t)}\!\gets\!\{i \in \set{I} \mid \vec{x}_{(k)}^{(t)}\!=\!Q_{\mat{\Xi}^{(t)}}(\vec{x}_{i}^{(t)})\}, \forall k \in \set{K}$
            \State Decoder: $\hat{\vec{y}}_{i}^{(t)}\!\gets\!D(\vec{x}_i^{(t)}), \forall i \in \set{I}$
            \State Compute loss $\Ls_{\text{CSG-AE}}(\Theta^{(t)}, \mat{\Xi}^{(t)}; \mat{A}, \set{Y_{I}})$
            \Statex \hspace{0.5em} \underline{\Comment{Backward pass}}
            \State Compute gradients $\nabla_{\Theta^{(t)}} \Ls_{\text{CSG-AE}}$ and $\nabla_{\mat{\Xi}^{(t)}} \Ls_{\text{CSG-AE}}$
            \Statex \hspace{0.5em} \underline{\Comment{Update parameters}}
            \State Update parameters $\Theta^{(t)}$ and $\mat{\Xi}^{(t)}$ by optimizer $\texttt{opt}$:
            \Statex \hspace{4em} $\Theta^{(t+1)}\!\gets\!\texttt{opt}(\Theta^{(t)}, \nabla_{\Theta^{(t)}} \Ls_{\text{CSG-AE}})$
            \Statex \hspace{4em} $\mat{\Xi}^{(t+1)}\!\gets\!\texttt{opt}(\mat{\Xi}^{(t)}, \nabla_{\mat{\Xi}^{(t)}} \Ls_{\text{CSG-AE}})$
        \EndFor
    \end{algorithmic}
\end{algorithm}

\subsection{Challenge 1: Low Codebook Utilization}
A codeword is deemed \textit{active} if at least one sample is assigned to it. Under the naive training approach in Algorithm~\ref{alg: CSG_AE_training_naive}, the model often suffers from a pathological state known as \textit{codebook collapse} \cite{kaiser2018fast, roy2018theory, huh2023straightening}, where only a small subset of codewords remain active, as shown by curve~1 in Fig.~\ref{fig:active-ratio}. While some codewords may reactivate during training, overall codebook utilization remains insufficient. This collapse results in degenerate solutions, constraining the information capacity of the codebook and hindering the model's ability to learn effective representations.

We attribute this phenomenon to a misalignment between embeddings and codewords, driven by two main factors: 
\paragraph{\textbf{Initialization Misalignment}}
When codebook vectors are randomly initialized to be far from the encoder-generated embeddings, as illustrated in Fig.~\ref{fig:tsne-visual}~(a) and (e), certain codewords remain unassigned and inactive. As a result, their loss terms then reduce to an MSE with the zero vector, pushing them toward the origin. 
\paragraph{\textbf{Embedding Drift}}
In early training, significant encoder parameter updates induce rapid embedding shifts (see Fig.~\ref{fig:tsne-visual}~(a) and (c)). This drift disrupts any transient codebook-activation patterns before they stabilize.

\subsection{Challenge 2: Dynamic Clustering Process}

Unlike conventional clustering where data points are static, our framework features a bidirectional dynamic clustering process: encoder parameters $\Theta$ continuously shift embeddings $\vec{x}_i$, while codebook parameters $\mat{\Xi}$ concurrently adjust grid centers $\vec{x}_{(k)}$. The naive synchronous parameter update strategy in Algorithm~\ref{alg: CSG_AE_training_naive} leads to two intertwined challenges:

\paragraph{\textbf{Assignment Update Hysteresis}}
The current updates of grid centers $\Delta \vec{x}_{(k)}^{(t)}$ depend on current embeddings $\vec{x}_i^{(t)}$,
    \begin{equation}
        \nabla_{\vec{x}_{(k)}^{(t)}} \Ls_{\text{CSG-AE}} = w_2 \nabla_{\vec{x}_{(k)}^{(t)}} \Ls_{2} \propto \left(\vec{x}_{(k)}^{(t)} - \vec{\mu}_{(k)}^{(t)} \right).
    \end{equation}
    which induces an update hysteresis. This temporal lag causes grid centers to continuously pursue outdated embeddings: as centers $\vec{x}_{(k)}^{(t)}$ adjust to $\vec{x}_{(k)}^{(t+1)}$ to match projected averages $\vec{\mu}_{(k)}^{(t)}$, the averages have already shifted to $\vec{\mu}_{(k)}^{(t+1)}$.
\paragraph{\textbf{Gradient Direction Conflicts}}
The updates of embeddings are influenced by the competing effects of reconstruction loss $\Ls_{1}$ and quantization loss $\Ls_{2}$:
\begin{equation}
    \nabla_{\vec{x}_i^{(t)}} \Ls_{\text{CSG-AE}} = w_1 \nabla_{\vec{x}_i^{(t)}} \Ls_{1} + w_2 \nabla_{\vec{x}_i^{(t)}} \Ls_{2}.
\end{equation}
If the angle between $\nabla_{\vec{x}_i^{(t)}} \Ls_{1}$ and $\nabla_{\vec{x}_i^{(t)}} \Ls_{2}$ exceeds $90^\circ$, their effects oppose each other.

In the presence of only quantization loss $\Ls_{2}$, the assignment update hysteresis would not pose a significant challenge, as the projected average centers and grid centers would naturally converge toward each other:
\begin{equation}
    \resizebox{0.89\linewidth}{!}{$\displaystyle
    \nabla_{\vec{x}_{i}^{(t)}} \Ls_{2} = -\frac{1}{{|\set{I}_{(k_i)}^{(t)}|}} \proj_{\set{N}_{(k_i)}^{(t)}} \left(\nabla_{\vec{x}_{(k_i)}^{(t)}} \Ls_{2}\right) \propto \vec{\mu}_{(k_i)}^{(t)}-\vec{x}_{(k_i)}^{(t)}.
    $}
\end{equation}
However, the reconstruction loss $\Ls_{1}$ introduces competing dynamics: it drives embeddings $\vec{x}_i$ toward positions that optimize reconstruction fidelity. When gradient conflicts emerge between these objectives, the assignment update hysteresis and gradient direction conflict effects intertwine in both temporal (assignment update hysteresis) and spatial (gradient direction conflict) dimensions, creating a complex optimization barrier for the training of the CSG-AE model.

\section{Clustering Metrics}
\label{appendix: clustering_metrics}

Consider a total of $I$ samples indexed by $\set{I}$, each associated with a true class label from $\set{C}$ and a predicted cluster label from $\set{K}$. For each $i \in \set{I}$, let $c_{i} \in \set{C}$ denote the true class and $k_{i} \in \set{K}$ the predicted cluster of the $i$-th sample. Define $n_{ck}$ as the number of samples with true label $c$ and predicted label $k$, that is, $n_{ck} = \sum_{i \in \set{I}} 1(c_{i} = c,\, k_{i} = k)$. The marginal counts are $n_c = \sum_k n_{ck}$ for class $c$ and $n_k = \sum_c n_{ck}$ for cluster $k$.

\subsection{Adjusted Rand Index}
The Adjusted Rand Index (ARI) \cite{hubert1985comparing} measures the similarity between two types of clustering labels, adjusting for chance. The ARI is defined as:
\begin{equation}
    \resizebox{0.89\linewidth}{!}{$\displaystyle
    \text{ARI} = \frac{\sum_{c,k} \binom{n_{ck}}{2} - \left[ \sum_{c} \binom{n_c}{2} \sum_{k} \binom{n_k}{2} \right] / \binom{I}{2}}{\frac{1}{2} \left[ \sum_{c} \binom{n_c}{2} + \sum_{k} \binom{n_k}{2} \right] - \left[ \sum_{c} \binom{n_c}{2} \sum_{k} \binom{n_k}{2} \right] / \binom{I}{2}},
    $}
\end{equation}
where \( \binom{x}{2} = \frac{x(x-1)}{2} \). The ARI ranges [-1, 1], with 1 meaning perfect agreement between the two clusterings, 0 denoting random agreement, and -1 indicating complete disagreement.

\subsection{Normalized Mutual Information}
Normalized Mutual Information (NMI) \cite{strehl2002cluster} quantifies the shared information between two clusterings, normalized to a scale from 0 to 1. The NMI is defined as:
\begin{equation}
    \text{NMI}(\set{C}, \set{K}) = \frac{\text{MI}(\set{C}, \set{K})}{\sqrt{H(\set{C}) H(\set{K})}},
\end{equation}
where \( H(\set{C}) = - \sum_{c} \frac{n_c}{I} \log \left( \frac{n_c}{I} \right) \) and \( H(\set{K}) = - \sum_{k} \frac{n_k}{I} \log \left( \frac{n_k}{I} \right) \) are the entropies of \( \set{C} \) and \( \set{K} \), respectively, and \( \text{MI}(\set{C}, \set{K}) \) is the Mutual Information (MI) between \( \set{C} \) and \( \set{K} \), which is defined as:
\begin{equation}
    \text{MI}(\set{C}, \set{K}) = \sum_{c,k} \frac{n_{ck}}{I} \log \left( \frac{I \cdot n_{ck}}{n_c n_k} \right),
\end{equation}
The NMI ranges [0, 1], with 1 denoting perfect correspondence between clusters and true classes, and 0 indicating complete independence between them.

\subsection{Completeness, Homogeneity, and V-Measure}
Completeness measures whether all samples of a given class are assigned to the same cluster. It is defined as:
\begin{equation}
    \text{Completeness}(\set{C}, \set{K}) = 1 - \frac{H(\set{K}|\set{C})}{H(\set{K})},
\end{equation}
where \( H(\set{K}|\set{C}) = - \sum_{c,k} \frac{n_{ck}}{n} \log \left( \frac{n_{ck}}{n_c} \right) \) is the conditional entropy of \( \set{K} \) given \( \set{C} \). Homogeneity assesses whether each cluster contains only members of a single class (\ie, purity):
\begin{equation}
    \text{Homogeneity}(\set{C}, \set{K}) = 1 - \frac{H(\set{C}|\set{K})}{H(\set{C})},
\end{equation}
similarly, \( H(\set{C}|\set{K}) = - \sum_{c,k} \frac{n_{ck}}{n} \log \left( \frac{n_{ck}}{n_k} \right) \) is the conditional entropy of \( \set{C} \) given \( \set{K} \). The V-Measure \cite{rosenberg2007v} is the harmonic mean of homogeneity and completeness:
\begin{equation}
    \text{V-Measure} = 2 \cdot \frac{\text{Homogeneity} \cdot \text{Completeness}}{\text{Homogeneity} + \text{Completeness}},
\end{equation}
All three metrics range [0, 1], with 1 indicating perfect agreement between the clustering and the ground truth labels and 0 indicating complete disagreement.


\ifCLASSOPTIONcaptionsoff
  \newpage
\fi
\bibliographystyle{IEEEtran}
\bibliography{ref}

\end{document}